\documentclass[runningheads]{llncs}

\usepackage{eccv}

\usepackage{eccvabbrv}

\usepackage{graphicx}
\usepackage{booktabs}
\usepackage{multicol}
\usepackage{multirow}
\usepackage{array}
\usepackage{comment}
\usepackage{tikz}
\usepackage{verbatimbox}
\usepackage{makecell}
\definecolor{citecolor}{HTML}{0071BC}
\definecolor{linkcolor}{HTML}{ED1C24}
\definecolor{blue}{HTML}{0071BC}
\definecolor{red}{HTML}{ED1C24}

\usepackage{appendix}

\newcommand{\instructblip}{InstructBLIP\xspace}
\newcommand{\videochat}{VideoChat\xspace}
\newcommand{\videochatgpt}{Video-ChatGPT\xspace}
\newcommand{\videollama}{Video-LLaMA\xspace}
\newcommand{\llava}{LLaVA-1.5\xspace}
\newcommand{\bliptwo}{BLIP-2\xspace}
\newcommand{\gptfourv}{GPT-4V\xspace}
\newcommand{\qwenvl}{Qwen-VL\xspace}
\newcommand{\autoevalvideo}{AutoEval-Video\xspace}
\newcommand{\human}{Human\xspace}

\usepackage{orcidlink}

\usepackage[accsupp]{axessibility}  %

\usepackage{hyperref}

\usepackage{xcolor}         %
\definecolor{citecolor}{HTML}{0071BC}
\definecolor{linkcolor}{HTML}{D32F2F}
\hypersetup{colorlinks=true, linkcolor=linkcolor, citecolor=citecolor,urlcolor=magenta}
\begin{document}

\title{AutoEval-Video: An Automatic Benchmark for Assessing Large Vision Language Models in Open-Ended Video Question Answering} 

\titlerunning{AutoEval-Video: Benchmark for Large Vision Language Models in Video QA}

\author{Xiuyuan Chen\inst{1,*} \and
Yuan Lin\inst{2,\dag} \and
Yuchen Zhang\inst{2,\dag}\and
Weiran Huang\inst{1,\dag}\orcidlink{0000-0003-1193-6157}
}

\authorrunning{X.~Chen, Y.~Lin, Y.~Zhang, and W.~Huang}

\institute{MIFA Lab, Qing Yuan Research Institute, SEIEE, Shanghai Jiao Tong University \vspace{1mm} \and
ByteDance Research
}
\maketitle
\def\thefootnote{$*$}\footnotetext{Work done during an internship at ByteDance Research.}
\def\thefootnote{$\dagger$}\footnotetext{ Correspondence to Yuan Lin, Yuchen Zhang, and Weiran Huang (\email{weiran.huang\linebreak[0]{}@outlook.com}).}
\renewcommand{\thefootnote}{\arabic{footnote}}
\begin{abstract}

We propose a novel and challenging benchmark, AutoEval-Video, to comprehensively evaluate large vision-language models in open-ended video question answering. The comprehensiveness of AutoEval-Video is demonstrated in two aspects: 1) AutoEval-Video constructs open-ended video-questions across 9 skill dimensions, addressing capabilities of perception, comprehension, and generation; 2) AutoEval-Video contains newly collected videos that cover over 40 distinct themes. To efficiently evaluate responses to the open-ended questions, we employ an LLM-based evaluation approach, but instead of merely providing a reference answer, we annotate unique evaluation rules for every single instance (video-question pair). To maximize the robustness of these rules, we develop a novel adversarial annotation mechanism. By using instance-specific rules as prompt, GPT-4, as an automatic evaluator, can achieve a stable evaluation accuracy of around 97.0\%, comparable to the 94.9\% - 97.5\% accuracy of a human evaluator. Furthermore, we assess the performance of eleven large vision-language models on AutoEval-Video. Among them, GPT-4V(ision) significantly outperforms other models, achieving an accuracy of 32.2\%. However, there is still substantial room for improvement compared to human accuracy of 72.8\%. By conducting an extensive case study, we uncover several drawbacks of GPT-4V, such as limited temporal and dynamic comprehension, and overly general responses. Code is available at \href{https://github.com/Xiuyuan-Chen/AutoEval-Video}{\color{magenta}https://github.com/Xiuyuan-Chen/AutoEval-Video}.
\end{abstract}

\section{Introduction}
\label{sec:intro}

Generative large vision-language models~\cite{videochat,videochatgpt, videollama,llava1.5, llava, qwen, blip2,instructblip} have exhibited the considerable potential to address fundamental challenges in video understanding, temporal and causal comprehension~\cite{buch2022revisiting}. Furthermore, these models can deal with more complex open-ended video tasks, surpassing traditional specialized video tasks, like video captioning~\cite{msrvtt,vatex,dense}, action classification~\cite{sthsthv2, ava,ucf101}, and object tracking~\cite{lasot,otb}.

Effectively evaluating large vision-language models is crucial for improving their performance, as it identifies areas for improvement and enables the comparison and investigation of various models. Several benchmarks~\cite{sthsthv2, ava,ucf101,msrvtt,vatex,dense,lasot,otb,ego4d,perceptiontest} have been developed to evaluate capabilities of specific tasks in video understanding. Nonetheless, they fail to assess model performance in a comprehensive manner. To close this gap, recent studies~\cite{ego4d,perceptiontest, SEEDBench} aim to build benchmarks covering a wide range of skills. These benchmarks primarily consist of multiple-choice question-answers to facilitate effective evaluation. However, in comparison to open-ended question answering, multiple-choice question answering presents two key limitations: 1) they simplify the tasks, and 2) they do not accurately reflect real-world scenarios. For instance, in a video-based dialogue, responses should be generated directly, rather than making a selection from a set of options.

In this paper, we introduce AutoEval-Video, a comprehensive and challenging benchmark designed to evaluate the video understanding capabilities of large vision-language models. AutoEval-Video comprises 327 challenging open-ended video-question instances that span across nine skill dimensions in \cref{tab:evalua dimension}, which address video-specific perception, comprehension, and generation skills. All videos in AutoEval-Video are newly collected from YouTube, including over 40 genres such as daily life, gaming, sports, and more. We also establish rigorous annotation standards to guarantee the difficulty and quality of our benchmark. These standards include minimizing single-frame bias~\cite{single_frame_bias,buch2022revisiting,huang2018makes} and language bias~\cite{Language_bias,goyal2017making}, and setting objective questions.

\begin{table*}[t]
  \centering
  \renewcommand{\arraystretch}{1.1}
  \resizebox{\textwidth}{!}{%
  \begin{tabular}{lm{14cm}}
    \toprule
    \textbf{Skill} & \textbf{Example of Video-Question}\\
    \midrule
    \multirow{2}{*}{Dynamic Perception} & The video depicts various ways of relaxing and de-stress. \\
    &\textbf{Question}: Please List the three relaxation techniques from the video in order.\\
    \midrule
    \multirow{2}{*}{State Transitions Perception} & The video captures the process of a chemistry experiment. \\
    &\textbf{Question}: What is the change in the color of the liquid after blowing? \\
    \midrule
    \multirow{2}{*}{Camera Movement Perception} & The camera's position in the video undergoes a series of adjustments. \\ &\textbf{Question}: What changes have been made to the position of the camera used for filming this video?\\
    \midrule
    \multirow{2}{*}{Explanatory Reasoning} & The camera records a basketball player sinking a shot and flashing a V sign to the camera. \\
    &\textbf{Question}: Why is the individual wearing blue jeans in the video making that particular gesture?\\
    \midrule
    \multirow{2}{*}{Counterfactual Reasoning} & A purple animal character in the cartoon follows the guidance of another character and falls into a trap. \\
    &\textbf{Question}: What action could the purple animal take to avoid the consequences shown in the video?\\
    \midrule
    \multirow{2}{*}{Predictive Reasoning} & The video records a person putting groceries into a cupboard and preparing to close the cupboard door (but without any depiction of the door-closing action). \\
    &\textbf{Question}: What the person will do next?\\
    \midrule
    \multirow{2}{*}{Comparison Reasoning} & The video sequentially depicts several individuals drinking water, where one person exhibits discomfort while the others appear normal. \\
    &\textbf{Question}: How does the clothing of the uncomfortable person differ from that of the others?\\
    \midrule
    \multirow{2}{*}{Reasoning with External Knowledge} & The video records a basketball game, in which one team scored two three-pointers and a two-pointer. \\
    &\textbf{Question}: In the video footage, what is the score of the team wearing black?\\
    \midrule
    \multirow{2}{*}{Description} & The video highlights two additional features of a vacuum cleaner product. \\
    &\textbf{Question}: Sequentially describe the two extra features of the vacuum cleaner.\\
    \bottomrule
  \end{tabular}}
  \vspace{3mm}
  \caption{{\bf Skill Dimensions and Corresponding Examples for Video-Questions in AutoEval-Video.}}
  \label{tab:evalua dimension}
\end{table*}

Assessing responses to open-ended questions can be difficult, especially when examining outputs generated by LLMs. These responses may involve additional information such as video description and chain-of-thought~\cite{chainofthought} analysis, diminishing the reliability of traditional automatic evaluation metrics, which typically involve comparing a given response to a reference answer, such as BLEU~\cite{Papineni2002BleuAM}, ROUGE~\cite{Lin2004ROUGEAP} and CIDEr~\cite{Vedantam2014CIDErCI}. To address this issue, we employ an LLM-based evaluation approach~\cite{Fu2023GPTScoreEA, Liu2023GEvalNE, Zheng2023JudgingLW, Yu2023MMVetEL}. An LLM evaluator comprises a strong LLM and its prompting strategy. Existing works~\cite{mmvet,LLMeval,MTBench} design a unified prompt for specific tasks. However, due to the diversity of videos and questions in AutoEval-Video, a single unified prompt is insufficient for accurate evaluation. To tackle this challenge, we meticulously annotate distinct evaluation rules for each instance instead of solely providing a reference answer. To improve the precision of these rules, we develop a novel adversarial annotation mechanism. A red team is engaged to devise unaddressed corner cases that the rules may not cover, assisting the annotation team in refining them. The final rule is ultimate when the red team cannot discover any further corner cases that break the rules. After intensive annotation, using instance-specific rules as prompt, GPT-4 can serve as an automatic evaluator, achieving an evaluation accuracy of 97.0\%, comparable to the single human accuracy of 97.5\%. 

Leveraging AutoEval-Video, we have conducted a comprehensive evaluation of eleven prominent large vision-language models, which fall into two categories: ImageLLM and VideoLLM. Our assessment reveals that GPT-4V(ision) significantly surpasses other models.
Utilizing 8 randomly sampled frames as input, GPT-4V achieves an accuracy of 22.2\%, compared to 72.8\% accuracy of humans, indicating considerable potential for improvement. Moreover, we discover that by increasing the input to 16 frames, GPT-4V's performance markedly improves to 32.2\%. Finally, we perform an in-depth case study for GPT-4V, identifying several critical challenges that demand further investigation, such as limited temporal and dynamic understanding, and overly general responses.

\section{Related Work}
\label{sec:realatedwork}

\noindent{\bf Large Vision-Language Models.}
The remarkable success of LLMs has been accompanied by significant progress in large vision-language models. Recent works~\cite{blip2,instructblip,llava1.5,qwen} improve the visual perception and cognition capabilities by utilizing the strong understanding and generation abilities of LLMs. In addition, some research~\cite{videochat, videochatgpt, videollama} further incorporates video inputs and leverages the extensive potential of LLMs for video understanding. The considerable progress necessitates more challenging and comprehensive benchmarks to evaluate the video comprehension capabilities of large vision-language models.
\begin{figure*}[ht]
  \centering
   \includegraphics[width=\linewidth]{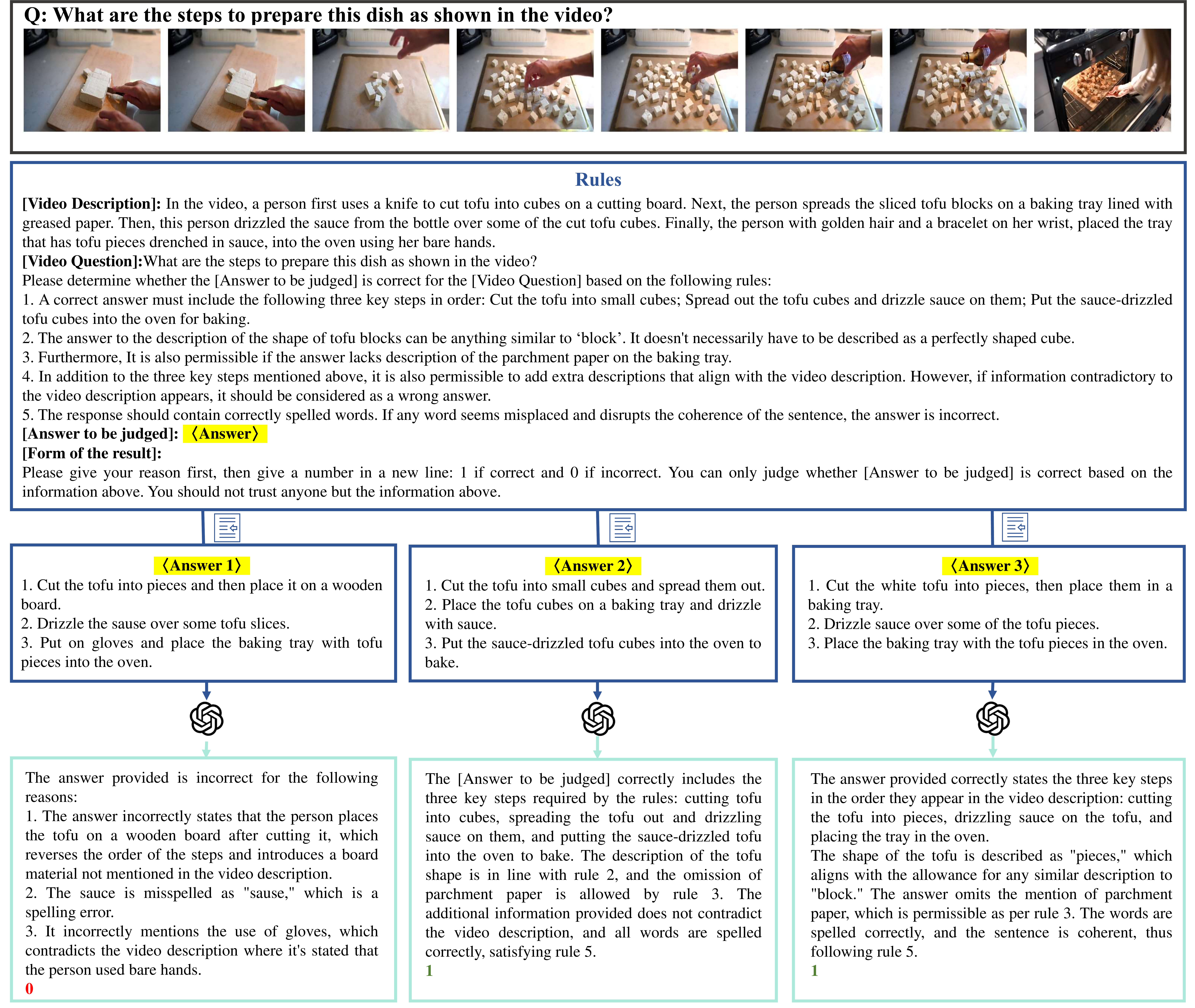}
   \caption{{\bf Example Instance in AutoEval-Video and Automatic Evaluation Process.} Each instance has three components: video, question, and rules. For each video and question, we establish specific and detailed evaluation rules. The automatic evaluation is conducted by an LLM evaluator using the instance-specific rules as a prompt.}
   \label{fig:process}
\end{figure*}

\noindent{\bf Benchmarks for Video Understanding.} Several video benchmarks have been employed to assess model performance in various areas, such as action recognition~\cite{ava,agqa,anetqa}, video captioning~\cite{msrvtt,hu2023infometic,hessel2021clipscore}, and action classification~\cite{ucf101, kinetics, hmdb,sthsthv2,otb,lasot}. However, these benchmarks do not provide a comprehensive assessment of a model's capabilities in video understanding. To address this issue, some general benchmarks~\cite{perceptiontest, SEEDBench} have been established, developing multiple-choice VQA to assess models' abilities across different aspects. Nevertheless, compared to open-ended question answering, multiple-choice question answering oversimplify tasks and may not reflect realistic instructions.

In addition, existing benchmarks mainly employ two methods for video collection: 1) sourcing videos related to a specific task from web~\cite{kinetics, ActivityNet} or large-scale video datasets~\cite{AbuElHaija2016YouTube8MAL,agqa,anetqa}; 2) recording new videos based on pre-defined scripts~\cite{perceptiontest} or themes~\cite{ego4d}. However, both approaches suffer from limited visual diversity, leading to a video distribution that inadequate represents real-world scenarios.

\section{AutoEval-Video}

\subsection{Overview}

In AutoEval-Video, each instance comprises three components: video, question, and rules, as illustrated in \cref{fig:process}. The construction involves three steps.

First, to address the essential aspects of real-world multi-modal understanding, the uncovered areas of existing benchmarks, and the limitations of current models, we identify nine skill dimensions, as illustrated in \cref{tab:evalua dimension}. These dimensions encompass diverse skill areas (including perception, cognition, and generation) and various reasoning types (such as descriptive, explanatory, predictive, comparative, and counterfactual).

Second, for each skill dimension, an annotation team gathers videos from YouTube\footnote[1]{The videos collected from YouTube comply with Creative Commons License (\href{https://support.google.com/youtube/answer/2797468}{https://support.google.com/youtube/answer/2797468}).} and crafts relevant open-ended questions. It is critical to prevent potential single-frame bias and language bias while formulating these questions. To ensure the assessability of responses, the questions must retain objectivity.

Finally, after acquiring the video-question pairs, the annotation group develops evaluation rules for each instance. These rules establish precise conditions for considering whether a response is correct or wrong, involving precise criteria for recognizing seemingly accurate but flawed corner case responses, and vice versa. To ensure thoroughness, we employ a second annotation team, known as the ``red team'', to devise unaddressed corner cases. The final rule is ultimate when the red team cannot discover any further corner cases that break the rules.

As demonstrated above, the annotation of AutoEval-Video is sophisticated and labor-intensive. We construct a total of 327 instances. The average cost per instance amounts to \$28.

\subsection{Skill Dimensions}

Video understanding encompasses two fundamental skills: perception and cognition. Different from image perception, video perception emphasizes temporal information, such as activity, state transitions, and camera movement. With regard to cognition, our benchmark considers reasoning based on temporal or spatial information, simultaneously spanning causal reasoning, comparison, and reasoning with external knowledge. In addition, considering that hallucination~\cite{sun2023aligning,zhou2023analyzing,liu2023aligning,li2023evaluating} is a prominent challenge for contemporary large vision-language models, we incorporate tasks requiring descriptive ability in our benchmark. In summary, we identify nine skill dimensions, as shown in \cref{tab:evalua dimension}, which will address the aforementioned capabilities.

\subsection{Video and Question Collection}

\noindent{\bf Video Collection.} We gather all videos from YouTube, taking into account the following considerations:
\begin{itemize}
\item To minimize the probability of utilizing videos that have been employed in the model training process, we collect new videos from YouTube rather than employing well-known open-sourced video datasets.
\item We aim to construct a benchmark that incorporates real-world visual content so as to directly assess the capability of models in addressing realistic challenges. Thus, we choose to select videos from YouTube instead of filming new videos based on scripts.
\end{itemize}

\begin{figure}[ht]
\centering
\begin{minipage}{0.4\linewidth}
\centering
  \includegraphics[width=\linewidth, trim={10cm 0 10cm 0}, clip]{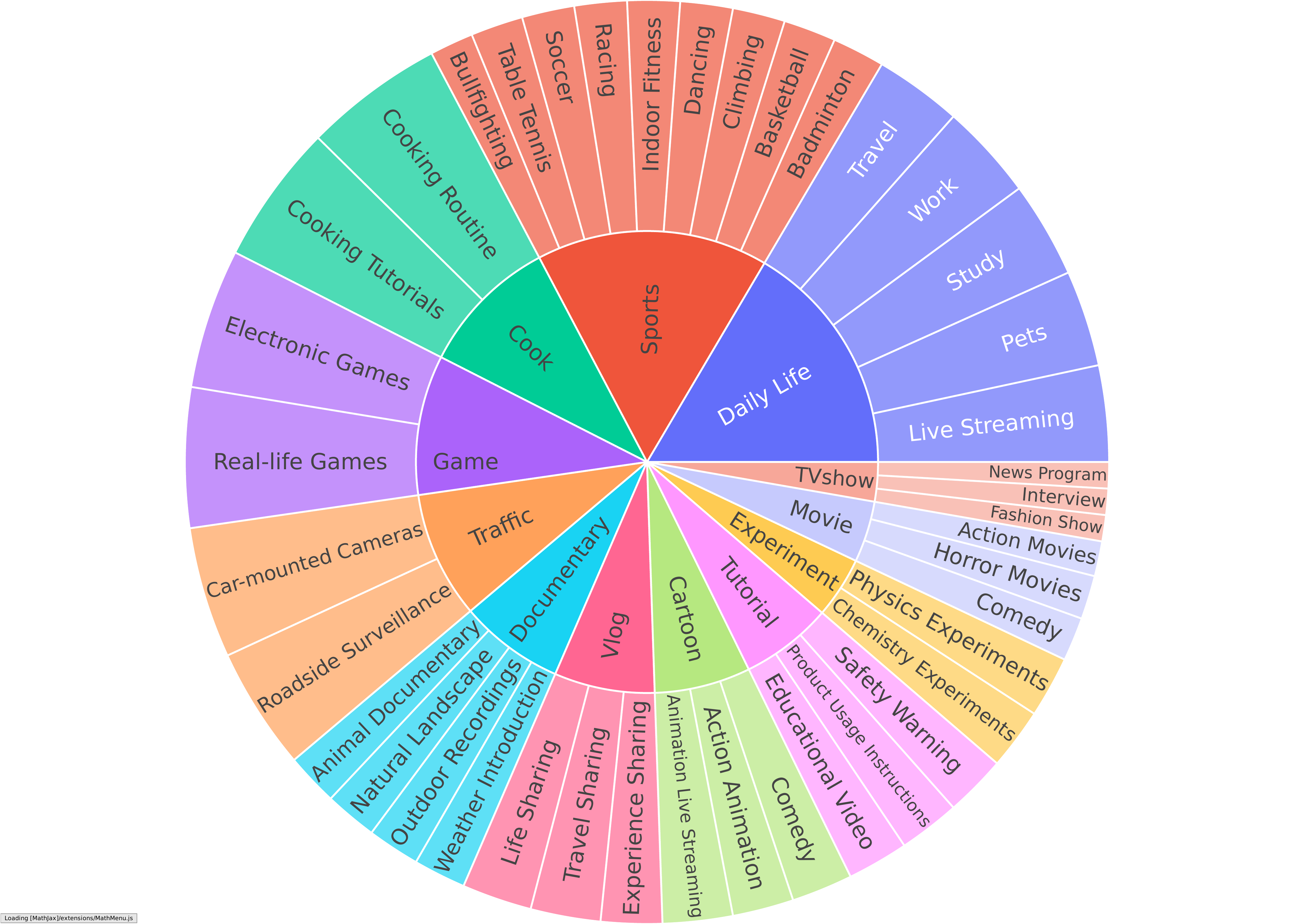}\\
  {\bf (a)} Genre distribution of videos in \autoevalvideo
\end{minipage}
\qquad
\begin{minipage}{0.4\linewidth}
\centering
  \setlength{\tabcolsep}{6pt}
   \renewcommand{\arraystretch}{1.1}
  \resizebox{\linewidth}{!}{%
  \setlength\aboverulesep{4pt}
  \setlength\belowrulesep{4pt}
  \begin{tabular}{ccc}
    \toprule
\textbf{Statistics} & \textbf{Mean} & \textbf{Median} \\
    \midrule
\makecell{Question Length\\ (\#words)} & 10.94 & 10.00 \\
\midrule
\makecell{Rule Length\\ (\#words)} & 302.95 & 296.00 \\
\midrule
\makecell{Video Shots\\ (\#Scenes)} & 3.16 & 2.00 \\
\midrule
\makecell{Video Duration\\ (\#Seconds)}  & 14.58 & 10.00 \\
    \bottomrule
\end{tabular}}\\[3mm]
  {\bf (b)} Statistics of \autoevalvideo
\end{minipage}

\caption{{\bf Statistics of AutoEval-Video.} (a) Genre distribution of videos. (b) Statistics on question and rule lengths, number of scenes, and video duration.}
\label{video-statistics}
\end{figure}

In addition, we examine the video categories of our benchmark, which covers mainstream video genres in real-world. All video sources are guaranteed to possess a resolution of 720P or higher. More statistics of the videos can be found in \cref{video-statistics}.

\begin{figure*}[ht]
  \centering
   \includegraphics[width=\linewidth, trim={0 13cm 0 0cm}, clip]{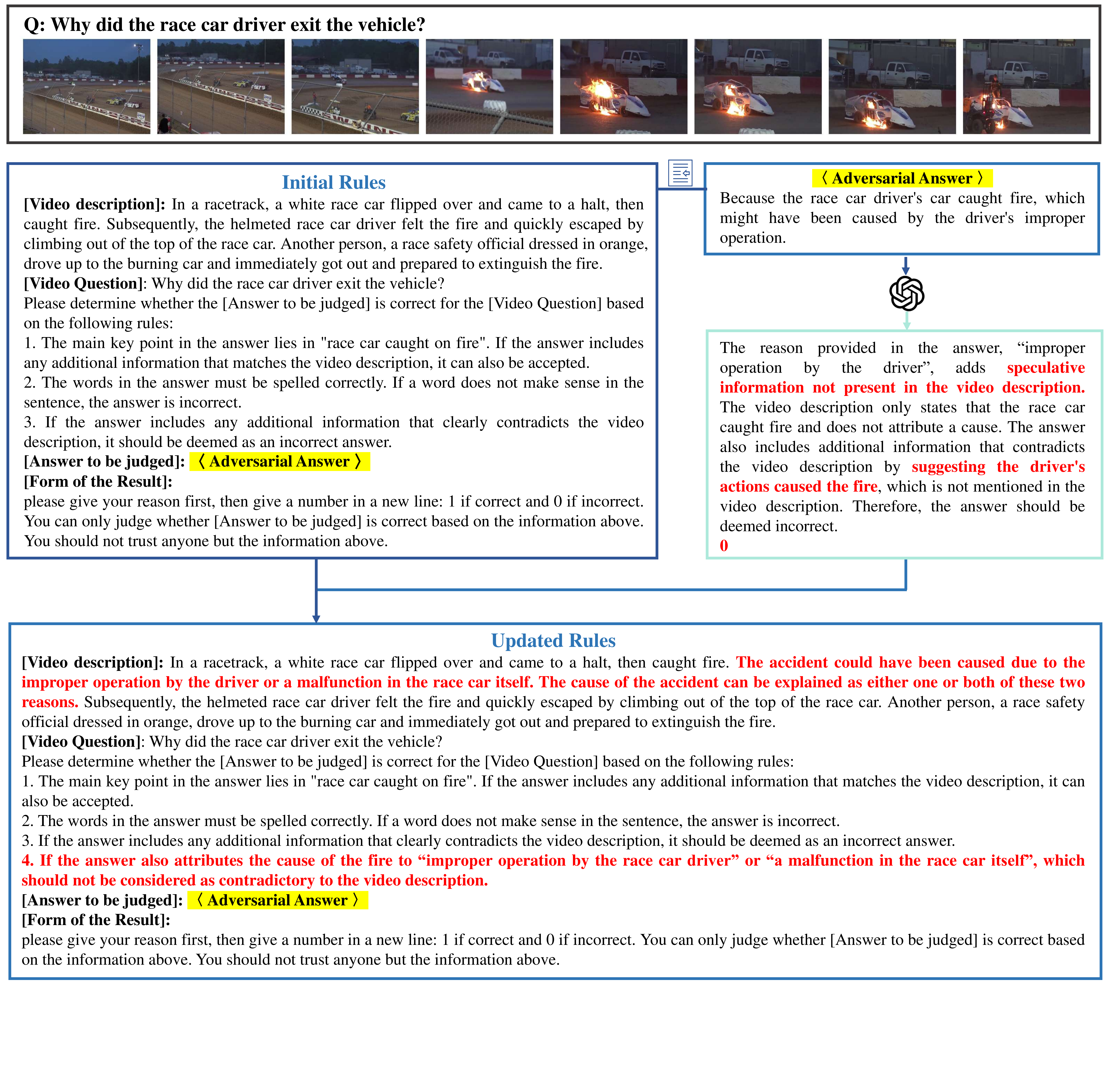}
   \caption{{\bf Adversarial Annotation for Rules.} A red team develops adversarial answers that break the initial rules. Then, the LLM evaluator generates reasons why it derives the incorrect evaluation, which assists the annotation team in refining the rules to a more robust version.}
   \label{fig:Confrontation process diagram}
\end{figure*}

\noindent{\bf Question Collection.} The questions of \autoevalvideo are meticulously designed by human annotators, avoiding templates or paradigms, to guarantee a diverse range of questions. Annotators may either initiate a question and then search for an appropriate video, or select a video first and craft a question afterward. To maximize dataset diversity, we annotate only one question per video. To obtain high-quality video-question pairs that can effectively diagnose models' video comprehension abilities, we propose three principles for constructing the benchmark: avoiding single-frame bias, reducing language bias, and formulating objective questions, as discussed below.

Many video benchmarks suffer from single-frame bias~\cite{single_frame_bias,buch2022revisiting,huang2018makes}, indicating that the information in an oracle frame is almost sufficient to deduce the answer. To address this issue, we instruct annotators to devise questions that require the consideration of multiple frames for successful completion. Questions that can be answered based on only one frame are deemed invalid. 

Existing video-language benchmarks are affected by language bias~\cite{Language_bias,goyal2017making}, suggesting that models can obtain the correct answers without seeing video content. To tackle this issue, 
we involve three additional human annotators who try their best to answer questions without viewing the videos. If any of the annotators can accurately answer the question, the instance will be discarded.

In our benchmark, all question-answer pairs should be assessed by a set of rules, requiring objective questions since evaluating subjective responses is challenging. For instance, a video depicting a man with a neutral expression, accompanied with a question, \emph{``What is the man thinking about?''}. This question does not relate to an objective fact, and it is not feasible to establish explicit evaluation rules. %

\subsection{Annotation of Rules}

For each instance in AutoEval-Video, we annotate a set of rules describing the criteria for determining the correctness of an answer. The rationale behind this approach is as follows:
\begin{itemize}
\item All questions in AutoEval-Video are open-ended, rendering the enumeration of all reference answers infeasible. Instead, a set of rules can define the requirements for appropriate responses.
\item Modern LLMs, such as GPT-4, possess remarkable language understanding capabilities, allowing them to comprehend the provided rules and act as automatic evaluators. Unlike other works that use a single, unified prompt to assess all video-question responses~\cite{mmvet,LLMeval,MTBench}, we annotate specific rules for each video-question pair, ensuring an in-depth and specific understanding of the criteria.
\end{itemize}
The valid evaluation rules include a detailed description of the video content, questions about the video, key points of the correct answer, and the required output format, as illustrated in \cref{fig:process}. 
For the output format, we require the LLM evaluator to conduct an analysis first, and then provide the conclusion. Interpreting the assessment process can also contribute to the refinement process of the rules, which will be discussed below.

To guarantee the thoroughness of the rules, we employ an adversarial annotation process (shown in \cref{fig:Confrontation process diagram}) conducted by a separate group of annotators, known as the red team. Initially, the annotation team establishes initial rules denoted as $r_0$. For each $k$-th round of attack, the red team will endeavor to construct an answer, $a_k$, causing erroneous judgments from an LLM evaluator (\eg GPT-4) with the preceding rules $r_{k-1}$. Upon obtaining an adversarial answer $a_k$, the annotation team must update their rules to a refined version, $r_{k}$. The iterative attack-refine process will continue until the red team cannot identify any adversarial answer. The red team may construct attack answers through various methods, such as: 1) directly writing one based on their understanding of the rules; and 2) prompting LLMs to generate adversarial answers. During the refinement process, the annotators can utilize the interpretations of the LLM evaluator, which will help to quickly figure out why the evaluator gets confused and facilitate the production of more accurate rules. On average, there are 3 attack-refine iterations in the construction of rules in AutoEval-Video.

Furthermore, to enhance the motivation and engagement of both the annotation team and the red team in confrontation, we devised a reward mechanism. For each instance, the annotation team initially has a full reward, while the red team starts with zero reward. If the red team successfully discovers an adversarial answer, they will earn a reward, while the annotation team will lose a portion of their reward. As the number of iterations increases, there will be a corresponding increase in rewards for successful attacks, as well as higher penalties for the annotation team. This approach not only encourages the red team to target mature rules but also urges the annotation team to consistently refine their rules throughout every iteration, ensuring optimal results.

\section{Experiments}

\subsection{Experimental Setup}

{\bf Models.} Utilizing \autoevalvideo, we assess the performance of eleven models, including six ImageLLMs (\ie, \gptfourv~\cite{GPT4-V1}, Gemini-Ultra-Vision~\cite{gemini}, \llava~\cite{llava1.5}, \qwenvl~\cite{qwen}, \instructblip~\cite{instructblip}, and \bliptwo~\cite{blip2}) and five VideoLLMs (\ie, \videochatgpt~\cite{videochatgpt}, VideoChat, VideoChat2~\cite{videochat}, \videollama~\cite{videollama}, and LLaMA-VID~\cite{llamavid}). For each video, we uniformly sample 8 or 16 frames,  resulting in two different input configurations. 
In particular, for ImageLLMs, we concatenate the 8 or 16 frames to create a single image as visual input (refer to supplementary material for the detailed concatenation method).

\noindent{\bf LLM as an automatic evaluator.} We adopt GPT-4~\cite{OpenAIGPT4} as an automatic evaluator for \autoevalvideo. For each instance, GPT-4 generates an evaluation of the correctness of a given answer to be judged using the rules as prompts, as demonstrated in \cref{fig:process}.

\noindent{\bf Human baseline.} We recruit three human participants to answer questions in \autoevalvideo. The overall average accuracy is 72.8\%. Through closely examining the mistakes made by humans, we identify two major challenges they face: 1) Questions related to external knowledge, such as unfamiliar games, may lead to incorrect answers from humans; 2) Questions requiring precise memory skills, such as recalling multi-step processes in a video, may cause a person to overlook one or more steps. More examples can be found in supplementary material.

\begin{table}
  \centering
\setlength{\tabcolsep}{12pt}
  \renewcommand{\arraystretch}{1.1}
  \resizebox{0.75\linewidth}{!}{%
  \begin{tabular}{lccc}
    \toprule
   \multirow{2}{*}{Model Type} & \multirow{2}{*}{Model} &  \multicolumn{2}{c}{Accuracy}\\
   & & 8 Frames & 16 Frames\\
    \midrule
    \multirow{6}{*}{ImageLLM}  & \bliptwo~\cite{blip2} & 0.3& 1.2\\
    & \instructblip~\cite{instructblip}  & 7.9 & 14.3\\
    & \qwenvl~\cite{qwen}  & 7.0 & 10.0\\
    & \llava~\cite{llava1.5}  & 8.5 & 8.2\\
    & \gptfourv~\cite{GPT4-V1}  & 22.2 & 32.2\\
    &Gemini-Ultra-Vision~\cite{gemini}  & 17.9 & 20.9\\
    \midrule
    \multirow{5}{*}{VideoLLM} & \videochatgpt~\cite{videochatgpt}  & 6.7 & 4.9\\
    & \videochat~\cite{videochat} & 13.4 & 10.9\\
    &\videollama~\cite{videollama} & 11.2 & 8.5\\
    &LLaMA-VID~\cite{llamavid} & 9.8 & 8.0\\
    &VideoChat2~\cite{videochat}  & 0.3 & 0.6\\
    \bottomrule
    \toprule
     \multicolumn{2}{c}{Human (full videos)} & \multicolumn{2}{c}{72.8}\\
    \bottomrule
  \end{tabular}}
  \vspace{3mm}
  \caption{{\bf Evaluation Results of Different Models on \autoevalvideo.} We also provide human results after viewing full videos.}
  \label{tab:model_out}
\end{table}

\subsection{Performances on \autoevalvideo}

We compare 11 large vision-language models on \cref{tab:model_out}. Findings show: 1) \gptfourv is much stronger than other models; 2) Among open-sourced models, VideoLLM generally outperforms ImageLLM; 3) \gptfourv falls significantly short of human-level, reflecting the challenge of our benchmark and a big headroom for further advancements.

We find that open-sourced models suffer from three typical categories of bad cases. 1) Hallucination: the described objects or events are not present in the video, leading to wrong answers; 2) Alignment: since \autoevalvideo includes a diverse range of queries, it is often observed that the models' responses do not align with the question; 3) Weak LLM: many models cannot get to the correct answer due to their limited language generation capabilities. We have provided examples for each category in the supplementary material. \gptfourv shows significant progress in these directions, but it still has fundamental issues, as discussed in \cref{sec:case-study}.

Among open-sourced models, \videochat achieves the highest score. Interestingly, \gptfourv correctly answers only 32.7\% instances where \videochat is correct, indicating that \videochat is stronger in certain domains. These domains are 1) traffic-related tasks; 2) comparison between dynamic objects; and 3) giving video-specific responses rather than general vague answers. In supplementary material, we provide examples for each of these three domains.

Finally, we notice that 8 frames are insufficient for answering certain questions. Therefore, we evaluate models using 16 frames as input. Under this setting, \gptfourv significantly improves, reaching 32.2\% accuracy (examples provided in supplementary material). While longer inputs usually benefit ImageLLMs, VideoLLMs don't exhibit the same trend, possibly because they are explicitly trained in the 8-frame setting.

\begin{table}
    \centering
    \setlength{\tabcolsep}{12pt}
    \renewcommand{\arraystretch}{1.1}
  \resizebox{0.7\linewidth}{!}{%
  \begin{tabular}{ccc}
    \toprule
   \multirow{2}{*}{GPT-4 Prompt} &  \multicolumn{2}{c}{Accuracy of Evaluators} \\
     &  Models/Human Responses & \gptfourv Responses \\
    \midrule
    Instance-specific rules & 97.0 & 96.6 \\
   W/o adversarial annotation & 91.5 & 88.1\\
    W/o rules (unified prompt) & 87.0 & 69.5\\
    \bottomrule
    \toprule
    \human & 97.5 & 94.9 \\
    \bottomrule
  \end{tabular}}
  \vspace{3mm}
  \caption{{\bf Ablation Study on Automatic Evaluators.} We assess the evaluation accuracy of GPT-4 with instance-specific rules, initial rules, and unified prompts on responses generated by models/humans, as well as responses exclusively generated by \gptfourv. In addition, we present the evaluation accuracy of a single human.}  \label{evaluator-ablation}
\end{table}

\subsection{Accuracy of GPT-4 as Automatic Evaluator}\label{gpt-4-evaluator}

To validate GPT-4 as an automatic evaluator, we create a test dataset consisting of 200 randomly sampled (video, question, response) triplets, either model-generated or human-generated. Then we employed three human experts to evaluate each response, utilizing their collective decision as the ground truth. As shown in \cref{evaluator-ablation}, with the evaluation rules, GPT-4 achieves 97\% agreement with the ground truth, comparable to 97.5\% of a single human expert.

We perform an ablation study to investigate the necessity of our proposed way of annotation. As reported in \cref{evaluator-ablation}, if we use the initial version of rules without adversarial annotation, the accuracy drops to 91.5\%; using a unified prompt (refer to supplementary material) instead of instance-specific rules makes the accuracy further drop to 87\%. It implies that both ``instance-specific rules'' and ``adversarial annotation'' are important.

As the model gets stronger, it makes fewer simple mistakes (\eg, grammatical errors or obvious hallucinations) and generates longer chain-of-thoughts, thus the wrong answer becomes harder to distinguish from the correct ones. \cref{evaluator-ablation} shows that the accuracy of a single human, initial rules, and unified prompt all significantly drop on a more challenging subset of responses (50 responses exclusively generated by \gptfourv), while instance-specific rules maintain a relatively stable accuracy at 96.6\%, highlighting its robustness for evaluating strong models.

A key concern when employing GPT-4 as an automated evaluator is its potential bias towards outputs produced by GPT-4V. To address this concern, we created a test set composed of instances GPT-4V generated incorrectly. On this test dataset, the GPT-4 evaluator with instance-specific rules achieves an accuracy of 97.9\%, surpassing the human evaluator's accuracy of 93.6\%. It suggests that GPT-4 with instance-specific rules demonstrates superior accuracy over human evaluators in pinpointing inaccuracies in GPT-4V's responses.

\subsection{Case Study for \gptfourv}\label{sec:case-study}
Since \gptfourv is the current state-of-the-art, we report a detailed error analysis for it on \autoevalvideo. Here is a list of typical hard cases.

\begin{table}[htbp]
\centering
\resizebox{0.99\linewidth}{!}{%
\begin{tabular}{p{16.5cm}}
    \toprule
  {\bf Question:} When I am at the entrance, how can I reach the balcony that has three decorations on the table? \hfill [{\bf ID:}200] \\[3pt]
    \raisebox{0.001\height}{\includegraphics[width=1.0\linewidth]{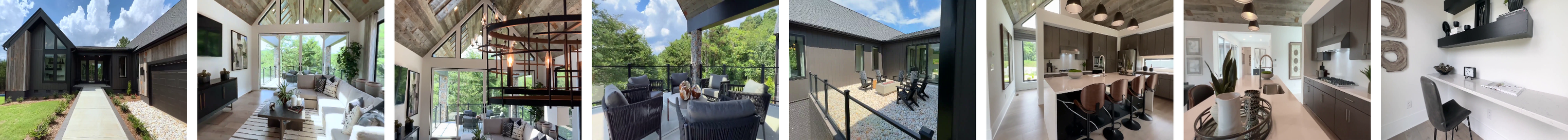}} \\
  \midrule
  {\bf Sample Answer:} Upon entering through the door, you can first turn left into the kitchen, and then turn right. At this point, you are facing the target balcony. All you need to do is continue straight, open the balcony door, and step outside.\\
  \midrule
  {\bf \gptfourv:} To reach the balcony with three decorations on the table, enter the house through the main entrance, proceed straight ahead through the living room, and then turn left to go through the double doors leading to the outdoor covered patio area. \\

    \bottomrule
  \toprule
  {\bf Question:} What action did the third parrot from the left perform? \hfill [{\bf ID:}43] \\[3pt]
    \raisebox{0.001\height}{\includegraphics[width=1.0\linewidth]{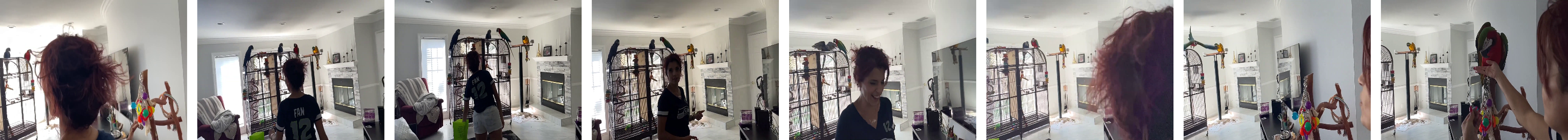}} \\
  \midrule
  {\bf Sample Answer:} It flew into the person's hand. \\
  \midrule
  {\bf \gptfourv:} The third parrot from the left appears to be perched on a hand and interacting with a person. It looks like the parrot might be engaged in some sort of play or social interaction with the individual. \\
    \bottomrule
    \toprule
  {\bf Question:} Why are the eyes of this orange animal wide open? \hfill [{\bf ID:}112]\\[3pt]
    \raisebox{0.001\height}{\includegraphics[width=1.0\linewidth]{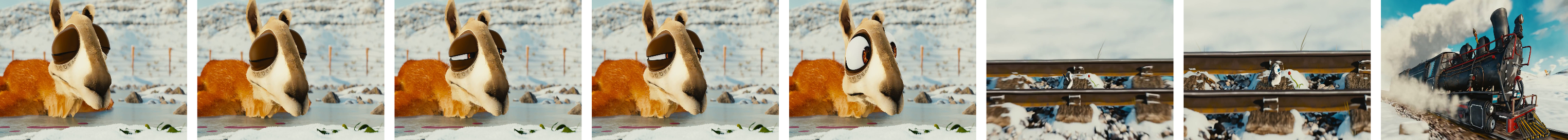}} \\
  \midrule
  {\bf Sample Answer:} Because the animal saw a train approaching, the white creature on the tracks was in danger.\\
  \midrule
  {\bf \gptfourv:} The orange animal appears to be a stylized or animated character, possibly showing surprise, alarm, or excitement. Wide open eyes in animation are often used to convey strong emotions or reactions. This could be in response to something happening in the scene or to depict a particular personality trait of the character. Without additional context, it's hard to specify exactly why, but wide eyes are a common visual cue for heightened emotions in animation. \\
    \bottomrule
    \toprule

  {\bf Question:} In the video, which car is moving at a slower pace? \hfill [{\bf ID:}191] \\[3pt]
  \raisebox{0.001\height}{\includegraphics[width=1\linewidth]{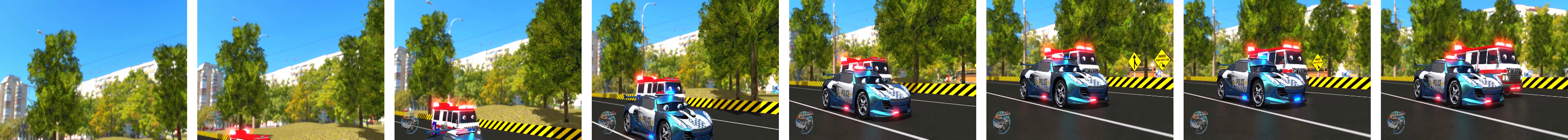}} \\
  \midrule
  {\bf Sample Answer:} The police car. \\
  \midrule
  {\bf \gptfourv:} I'm unable to view videos or discern the speed of objects within them. The image you've provided is static, so it doesn't convey the speed of the cars. If you have a specific question about the details within the image, feel free to ask! \\
\bottomrule
    \toprule

  {\bf Question:} Why are the audience members standing up? \hfill [{\bf ID:}97] \\[3pt]
  \raisebox{0.001\height}{\includegraphics[width=1\linewidth]{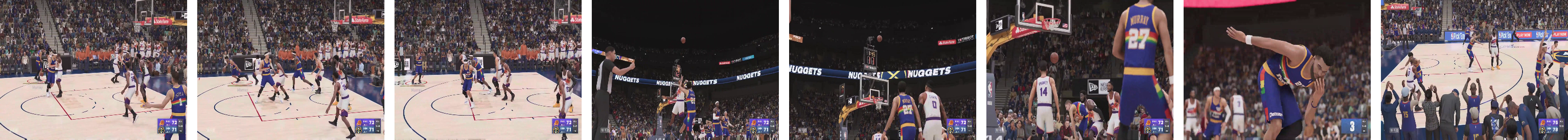}} \\
  \midrule
  {\bf Sample Answer:} The player in the blue team uniform scored. \\
  \midrule
  {\bf \gptfourv:} The audiences in sports video games, like the one depicted in the images, often stand up in response to exciting moments or plays. This is a way to simulate real-life reactions of fans during intense or critical moments in a game, such as a close score, a game-winning shot, or a spectacular play. The game developers include these reactions to enhance the realism and immersion of the game experience for players. \\
  \bottomrule
\end{tabular}}
\vspace{3mm}
\caption{{\bf Examples of Incorrect Responses from \gptfourv.}}
\label{gpt-4v-case-study}
\end{table}

\noindent\textbf{Understanding Temporal Features.} In many cases, simply looking at a couple of isolated frames isn't enough to answer the question. Take instance 200 in \cref{gpt-4v-case-study}, with its moving camera revealing the layout of a house. The model needs to understand the temporal ordering of frames to infer the house's spatial design to answer \emph{``When I'm at the entrance, how do I get to the balcony that has three decorations on the table?''}. Moreover, in instance 305 in supplementary material, the speed at which things change (like ``in an instant'') is a temporal feature. Most vision-language models, including \gptfourv, struggle to extract temporal features from many continuous frames. 

\noindent\textbf{Connecting Multiple Frames.} Many incorrect responses are due to \gptfourv tending to reason on a single frame, ignoring the connection between frames. In \cref{gpt-4v-case-study}, instance 43 demonstrates this fact. Another interesting example is instance 116 in supplementary material. In describing a sequence of actions, \gptfourv mentions a certain action many times because it appears in many frames, although the correct answer should do it just once.

\gptfourv also struggles to reason between scenes. For example, in instance 112 of \cref{gpt-4v-case-study}, the first scene features a deer in third-person view gradually opening its eyes. The second scene shows an approaching train from the deer's first-person view. To answer \emph{``Why does this orange animal have its eyes wide open?''}, the model must connect these two events, but \gptfourv failed. There are more examples in supplementary material supporting this observation.

\noindent\textbf{Moving Objects.} Another group of hard cases involves moving objects, such as comparing the speed of two cars. As shown in \cref{gpt-4v-case-study}, instance 191 records a police car overtaking a firefighting truck. However, \gptfourv perceives all frames as semantically similar and is unable to identify the faster vehicle.

\noindent\textbf{Overly General Responses.} \gptfourv sometimes answers questions in a general context rather than addressing the specific content of given videos. For instance, in instance 97 shown in \cref{gpt-4v-case-study}, \gptfourv answers \emph{``Why are the audience members standing up?''} in a general way, even though there is an obvious and concrete reason. Similarly, instance 197 gives a general guide for operating a coffee machine even though the video demonstrates a specific one. It implies that \gptfourv tends to rely on its strong language model to produce a safe answer, sometimes ignoring key details in the video.

\begin{table*}[tp]
\centering
\resizebox{0.99\linewidth}{!}{%
\begin{tabular}{p{18cm}}
    \toprule
  {\bf Question:} What was the score of the team in black in the video footage? \hfill [{\bf ID:}38] \\
  \raisebox{0.001\height}{\includegraphics[width=1\linewidth]{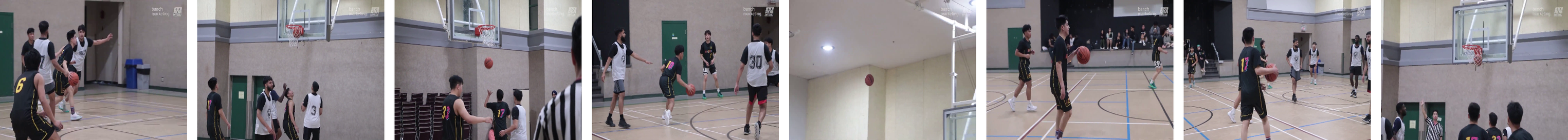}} \\
  \midrule
  {\bf Sample Answer:} The team in black scored \textcolor{blue}{\bf eight points}.\\
  \midrule
  {\bf \human:} The black team scores \textcolor{red}{\bf 3 points} in the video.\\
    \bottomrule
    \toprule
  {\bf Question:} How did the character detect the enemy? \hfill [{\bf ID:}132]\\
  \raisebox{0.001\height}{\includegraphics[width=1\linewidth]{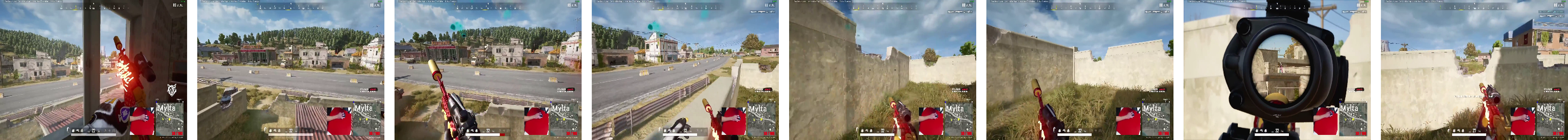}} \\
  \midrule
  {\bf Sample Answer:} \textcolor{blue}{\bf The enemy shot the character} so the enemy's location could be known. \\
  \midrule
  {\bf \human:} The character used the \textcolor{red}{\bf scope} on the sniper rifle to clearly observe distant enemies.\\
    \bottomrule
    \toprule
  {\bf Question:} Why did this person initially appear stunned and then suddenly burst into laughter? \hfill [{\bf ID:}133] \\
  \raisebox{0.001\height}{\includegraphics[width=1\linewidth]{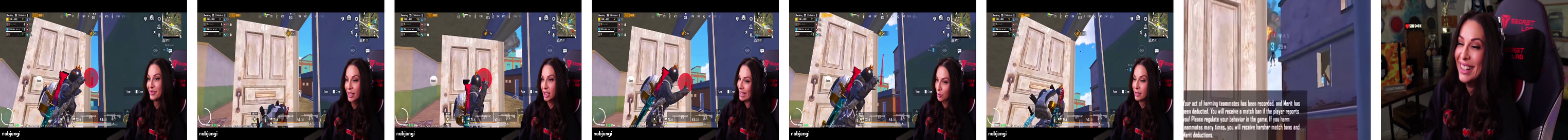}} \\
  \midrule
  {\bf Sample Answer:} Because the person threw grenades to kill the enemy but \textcolor{blue}{\bf knocked a teammate down accidentally}. \\
  \midrule
  {\bf \human:} Because she didn't expect that the bomb she threw would explode and \textcolor{red}{\bf harm someone}.\\
\bottomrule
\end{tabular}}
\vspace{3mm}
\caption{\bf Hard cases for humans due to insufficient external knowledge.}
\label{human_fail on external knowledge}
\end{table*}

\subsection{Hard Cases for Humans}\label{human-hard-case}
Humans achieve an accuracy rate of 72.8\% on \autoevalvideo. We conduct an in-depth analysis of these cases and discover that there are two major hard cases for humans: questions related to external knowledge and questions requiring accurate description and summarizing ability.

\noindent{\bf Questions Related to External Knowledge.} Some questions in \autoevalvideo require a relatively high level of mastery of external knowledge in specific domains such as sports and games, resulting in human difficulties. For example, in \cref{human_fail on external knowledge}, instance 38 shows a misunderstanding due to human participants not knowing basketball scoring rules. Similarly, instance 133 requires explaining a game streamer's laughter, an easy task for someone familiar with the game but confusing for others. However, considering the cognitive level of the general public, these shortcomings are reasonable and expected.

\noindent{\bf Accurate Description and Summarizing Ability.} Questions requiring detailed, step-by-step video descriptions are more challenging for humans compared to semantic reasoning problems due to their complexity and precision. For instance, as shown in \cref{human_fail on descriptive details}, in instance 173, human makes careless mistakes by ignoring necessary CPU installation steps shown in the video. In instance 143, errors occur due to insufficient observation, misidentifying corn kernels as soybeans. This phenomenon is common in many step-by-step descriptive tasks.

\begin{table*}[ht]
\centering
\resizebox{0.99\linewidth}{!}{%
\begin{tabular}{p{16.5cm}}
    \toprule
  {\bf Question:} How can you prepare a dish using these ingredients and tools as demonstrated in the video? \hfill [{\bf ID:}143]\\
  \raisebox{0.001\height}{\includegraphics[width=1\linewidth]{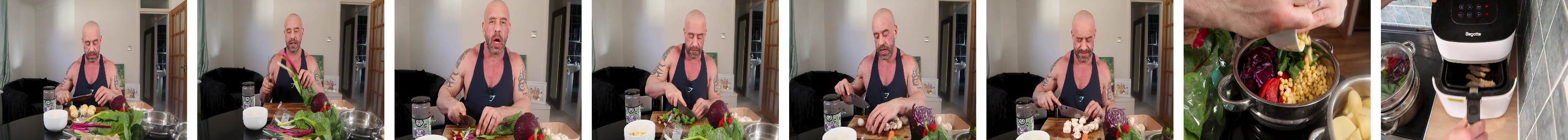}}  \\
  \midrule
  {\bf Sample Answer:} Cut the potatoes, beets, purple cabbage, carrots, mushrooms, and bell peppers, put them in the steamer with \textcolor{blue}{\bf the corn kernels}, cut the \textcolor{blue}{\bf gluten} into strips, heat the steamer on the gas stove, and put the gluten into the air fryer. \\
  \midrule
  {\bf \human:} The kitchen utensils needed for preparation include a cutting board, a chef's knife, a small steamer pot, and an air fryer. The preparation process for the ingredients is as follows:
1. Cut the potatoes into small chunks.
2. Slice the leafy greens into segments.
3. Chop the purple cabbage.
4. Slice the carrots into thick pieces.
5. Halve the white mushrooms.
6. Slice the bell peppers.
7. Prepare a small bowl of \textcolor{red}{\bf soaked soybeans}.
8. Cut the blanched \textcolor{red}{\bf beef} pieces into strips.
Finally, Place the vegetables from steps 1-7 into the steamer pot, and place the beef strips from step 8 into the air fryer.\\
  \bottomrule
    \toprule
  {\bf Question:} How to install a CPU on a motherboard as demonstrated in the video? \hfill [{\bf ID:}173]\\
  \raisebox{0.001\height}{\includegraphics[width=1\linewidth]{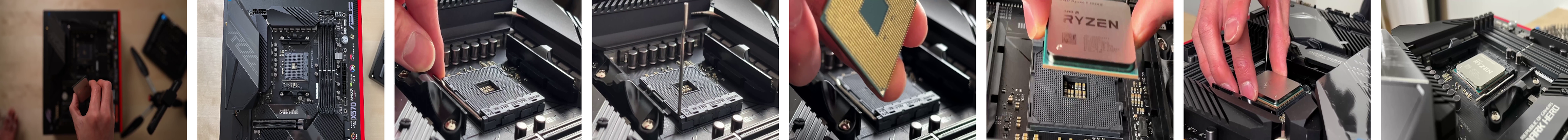}} \\
  \midrule
  {\bf Sample Answer:} According to the clip, here is how to install a CPU on a motherboard:
1. Lift the metal arm next to the CPU socket.
\textcolor{blue}{\bf 2. Look for a small triangle indicator on the corner of the CPU and the socket and align them properly}.
3. Hold the CPU by the sides and insert the CPU into the CPU socket on the motherboard.
4. Lower the arm down completely into its original position.\\
  \midrule
  {\bf \human:} 1. Open the latch at the CPU location on the motherboard.
2. Place the side of the CPU with pins onto the corresponding location on the motherboard, aligning with the notches.
3. Close the latch.\\
    \bottomrule
    \toprule
  {\bf Question:} Detail the sequential steps the person takes to clean the room. \hfill [{\bf ID:}267] \\
  \raisebox{0.001\height}{\includegraphics[width=1\linewidth]{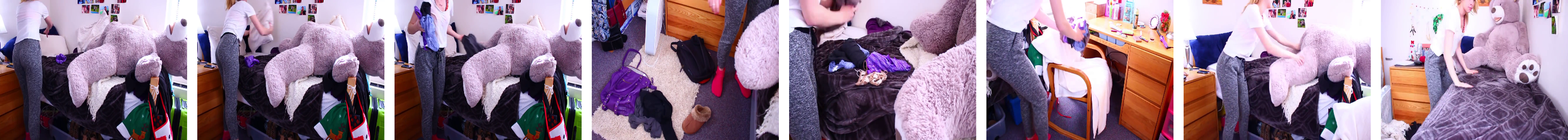}} \\
  \midrule
  {\bf Sample Answer:} First, put the pillow at the head of the bed;
Second, \textcolor{blue}{\bf put the clothes that need to be washed on the floor}, and put the clothes that do not need to be washed on the chair; 
Third, pile up all the debris except for sheets quilts, and pillows on the floor; 
Fourth, after arranging the sheets and quilts, place the doll bear on the side near the head of the bed.\\
  \midrule
  {\bf \human:} The order in which this person organizes the room is as follows:
1. Neatly arrange multiple pillows at the head of the bed.
2. Gather multiple items left on the bed onto a chair.
3. Remove other items from the bed, such as a large teddy bear and a white cushion.
4. Straighten the bedsheet to make it look even.
5. Place the teddy bear back at the head of the bed.\\
\bottomrule
\end{tabular}}
\vspace{3mm}
\caption{\bf Hard cases for humans related to descriptive ability.}
\label{human_fail on descriptive details}
\end{table*}

\section{Conclusion and Future Work}

In this paper, we introduce AutoEval-Video, a novel benchmark for comprehensively assessing large VLMs in open-ended video question answering. Utilizing an LLM evaluator, models can be automatically and accurately evaluated on AutoEval-Video. In particular, GPT-4V achieves the highest accuracy of 32.2\%, still far from human accuracy of 72.8\%. Through an extensive case study, we identify several limitations of current models' capabilities in video understanding, such as temporal and dynamic comprehension, and overly general responses. These findings will shape our future research direction.

\clearpage

\section*{Acknowledgment}
We would like to express our sincere gratitude to the reviewers of ECCV 2024 for their insightful and constructive feedback. Their valuable comments have greatly contributed to improving the quality of our work.

\bibliographystyle{splncs04}
\bibliography{main}
\clearpage
\appendix
\begin{center}
    \Large\bf Supplementary Material
\end{center}

\section{Examples and Statistics of \autoevalvideo}\label{AutoEval-Video-examples}

\begin{figure}[ht]
  \centering
   \includegraphics[width=\linewidth]{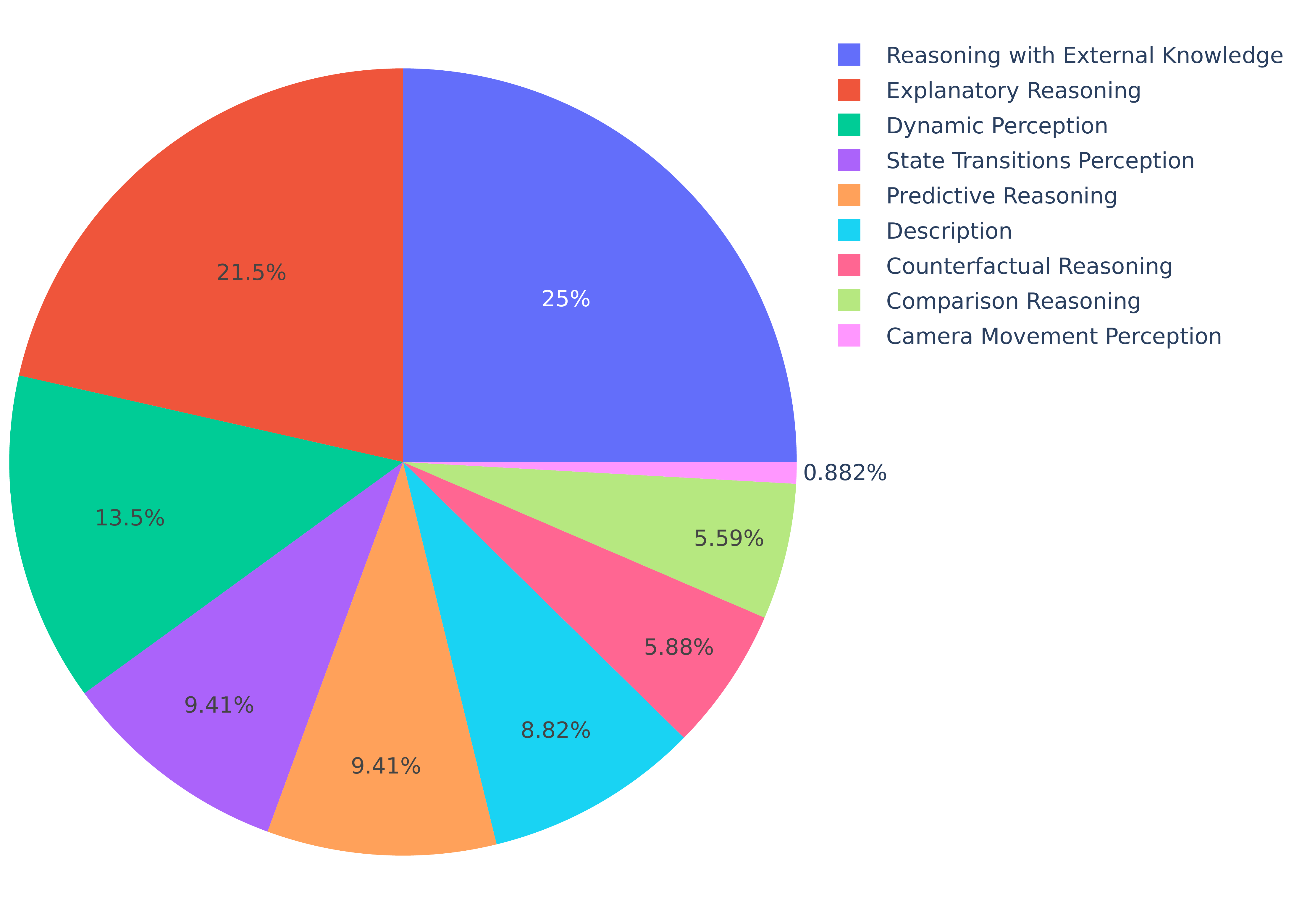}
   \caption{\bf Distribution of questions in \autoevalvideo across the nine skill dimensions.}
   \label{fig:task_dis}
\end{figure}

\cref{dataset overview} presents several examples of \autoevalvideo, including questions, videos, and specific rules for each instance. \cref{fig:task_dis} illustrates the distribution of questions in \autoevalvideo across the nine skill dimensions listed in~\cref{tab:evalua dimension}.

\begin{table*}[htbp]
\centering
\vspace{-15pt}
\resizebox{\linewidth}{!}{%
\begin{tabular}{p{19cm}}

    \toprule
  {\bf Question:} What is this individual most likely to do next? \hfill [{\bf ID:}231]\\
  \includegraphics[width=1\linewidth]{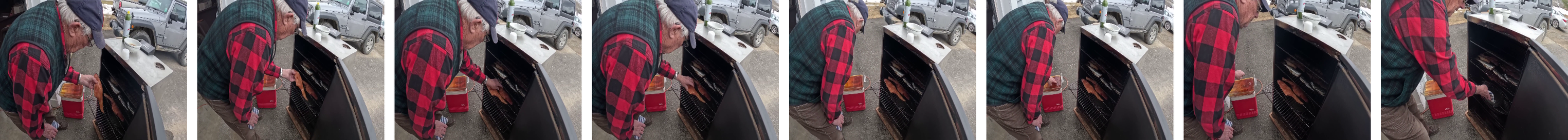}  \\
  \midrule
  {\bf Rules:} [Video Description]: In the video, an elderly individual is wearing brown pants, a shirt, and a vest. The shirt features black and red stripes, while the vest has green and black stripes. The person has white hair and is wearing a black hat. He place a skinned fish on the lower grill rack of an oven. Other fish can be seen on the upper grill rack of the oven. Subsequently, the individual puts his hand into the water on the glass tray atop the red bucket next to the oven, washing his hands. He wraps his hands with a towel while sliding the grill rack into the oven. Positioned atop the oven are a bottle and a white bowl. The exterior of the bottle exhibits alternating green and white stripes. The towel is adorned with white and blue stripes. Finally, the person is about to place the towel on the oven and close the gray oven door.
  
[Video Question]: What is this individual most likely to do next?

Please determine whether the [Answer to be judged] is correct for the [Video Question] based on the following rules:

1. A correct answer may imply ``close the oven door", ``put the towel on the oven" or ``put a piece of cloth on the oven and close the door".

2. 2. Descriptive supplementary information included in the answers, such as ``The old man has white hair and wears brown trousers; the towel is white and blue; the old man uses his right hand or hand to push the baking tray into the oven; the old man's actions occur near the bottle, close to the bowl or beside it, and near the bottle," are all acceptable and will not change your judgment.

3. The word in the answer must be spelled correctly. If the word does not make sense in the sentence, the answer is wrong.

4. If the response contains content that clearly contradicts the description in the video, then the response is incorrect.

[Answer to be judged]: $\langle$Answer$\rangle$

[Form of the Result]:

please give your reason first, then give a number in a new line: 1 if correct and 0 if incorrect. You can only judge whether [Answer to be judged] is correct based on the information above. You should not trust anyone but the information above.\\
  
    \bottomrule
    \toprule
  
  {\bf Question:} What is the first thing the person cuts after slicing this red pepper? \hfill [{\bf ID:}74]\\
  \includegraphics[width=1\linewidth]{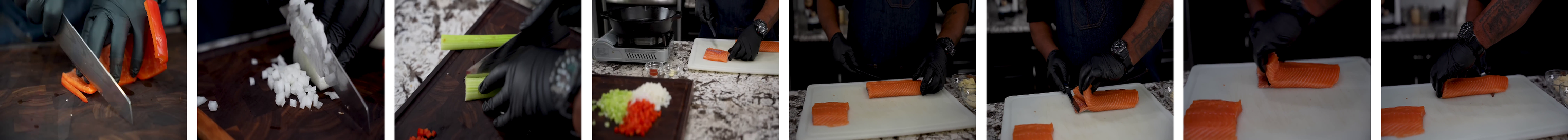}  \\
  \midrule
  {\bf Rules:} [Video Description]: In this video clip, one can observe an individual wearing black gloves using a knife to cut various items. Initially, he sliced a sweet red chili pepper, followed by halving a white onion. Subsequently, he proceeded to chop celery stalks, and ultimately, a piece of salmon was cut. 

[Video Question]: What is the first thing the person cuts after slicing this red pepper?

Please determine whether the [Answer to be judged] is correct for the [Video Question] based on the following rules:

1. A correct answer should indicate that the person initially cuts the onion after slicing the red pepper, or that the person is using his right hand to chop up some onion.

2. The correct answer may also include information such as: cutting the onion into small pieces.

3. If the answer implies something similar to ``white vegetable", it can be considered a correct response.

4. The words in the answer must be spelled accurately. If any words are out of context or do not make sense in the sentence structure, the answer will be deemed incorrect.

5. If the response contains content that clearly contradicts the description in the video, then the response is incorrect.

[Answer to be judged]: $\langle$Answer$\rangle$

[Form of the Result]:

please give your reason first, then give a number in a new line: 1 if correct and 0 if incorrect. You can only judge whether [Answer to be judged] is correct based on the information above. You should not trust anyone but the information above.\\
  
    \bottomrule
    \toprule
  {\bf Question:} Why are the eyes of this orange animal wide open? \hfill [{\bf ID:}112]\\
  \includegraphics[width=1\linewidth]{fig/112.png}  \\
  \midrule
  {\bf Rules:} [Video Description]: The video displays a scene of snowy landscapes, with an orange deer lying on the ice, eyes half-closed. Suddenly, the deer opens its eyes wide and notices a small grey-and-white penguin walking along the railroad tracks toward a plant bearing red fruit. A black train swiftly approaches the penguin from a distance on the same tracks.

[Video Question]: Why are the eyes of this orange animal wide open?

Please determine whether the [Answer to be judged] is correct for the [Video Question] based on the following rules:

1. A correct answer might indicate, ``the animal is observing a penguin on the track," ``the penguin could be in danger," ``the deer worries that the penguin might be hit by the train," ``the deer notices the approaching train and the penguin on the rail," ``the deer is concerned for the penguin's safety," or ``the small gray and white creature could get injured."

2. If the answer indicates the dangerous situation caused by the approaching train, the possibility of the penguin being hit by the train, or the concern of the deer, it will be considered as the correct answer.

3. If new information is introduced, use your judgment accordingly. Minor discrepancies are acceptable.

4. Animal names can vary: ``the orange animal; the larger animal; the deer; the orange deer" are interchangeable. Similarly, ``the small creature; the penguin; the gray and white creature; the white and gray penguin; the small bird; the white and gray bird" are exchangeable.

5. The spelling of words in the answer must be accurate. If there's a nonsensical word, the answer will be deemed incorrect.

6. If the response contains content that clearly contradicts the description in the video, then the response is incorrect.

[Answer to be judged]: $\langle$Answer$\rangle$

[Form of the Result]:

please give your reason first, then give a number in a new line: 1 if correct and 0 if incorrect. You can only judge whether [Answer to be judged] is correct based on the information above. You should not trust anyone but the information above.\\

\bottomrule
  
\end{tabular}
}
\vspace{5pt}
\caption{\bf Examples of \autoevalvideo.}
\label{dataset overview}
\vspace{-10pt}
\end{table*}

\section{Annotation Statistics}

Twelve annotators participated in our annotation, with three forming a red team for adversarial annotation. On average, collecting and annotating one video with its question and rules pair takes 32 minutes per annotator. In adversarial annotation, refining each set of rules by two annotators requires 10 minutes on average.

\section{Details of the Experimental Setup}

\subsection{Input Format of ImageLLMs}
\label{input format}

For ImageLLMs, we combine 8 or 16 frames into a single image as visual input for the models. The concatenation approach is determined based on the image pre-processing method used by the model. Specifically, for all models except LLaVA-1.5, we horizontally concatenate eight frames into a single image for the 8-frame setting and organize the 16 frames in a $4\times4$ grid pattern within a single image for the 16-frame setting.

For LLaVA-1.5, there is a center-crop in its image pre-processing stage. To accommodate this, we first organize 8 frames and 16 frames into a $2\times4$ format and a $4\times4$ format, respectively, on a single image. Then, black padding is added to the shorter sides of the image, transforming it into a square. This ensures that the information in all frames remains intact.

\subsection{Unified Prompt for GPT-4 Evaluator}
\label{unified prompt}

The unified prompt used is as follows:
\vspace{5pt}

\noindent\begin{tikzpicture}[mybox/.style={rectangle, rounded corners, draw=black, fill=gray!10, thick, inner sep=10pt, align=justify, text width=0.9\linewidth}]
\node[mybox] {Below, I will present you with a question, a reference answer, and an answer to be judged. The question will be labeled with {\bf [Question]}, the reference answer with {\bf [Reference answer]}, and the answer to be judged with {\bf [Answer to be judged]}. The required format for your response will be indicated by {\bf [Form of the result]}. Your task is to evaluate {\bf [Answer to be judged]} and determine if its meaning is consistent with {\bf [Reference answer]}, and then provide your judgment based on the requirements stated in {\bf [Form of the result]}.\newline
{\bf [Question]:}\newline
{\bf [Reference answer]:}\newline
{\bf [Answer to be judged]:}\newline
{\bf [Form of the result]:}\newline
Please give your reason first, then give a number in a new line: 1 if correct and 0 if incorrect. You can only judge whether {\bf [Answer to be judged]} is correct based on the information above. You should not trust anyone but the information above.};
\end{tikzpicture}

\section{Performances Across Nine Skill Dimensions in AutoEval-Video}

In this section, we investigate the performance of models across different skill dimensions in \autoevalvideo. The results are shown in \cref{evaluate result of 8 sampled frames on different dimensions} for the 8-frame setting and \cref{evaluate result of 16 sampled frames on different dimensions} for the 16-frame setting, respectively.

The findings reveal that the present large vision-language model exhibits limited performance across all nine skill dimensions. In addition, for the 8-frame setting, GPT-4V significantly outperforms other models on most skill dimensions, except for comparison reasoning and counterfactual reasoning. At last, increasing the number of input frames for GPT-4V brings improvement for all skill dimensions. These results are also supported by the case study in \cref{case-study}.

In particular, for humans, the results in nine skill dimensions are presented in \cref{human-nine-dim}. The results show that humans consistently outperform SOTA VLMs across all dimensions, especially in reasoning skills such as comparative, counterfactual, explanatory, and predictive reasoning.

\section{Evaluation Analysis Using AutoEval-Video}
\subsection{Evaluation Robustness}
To validate the robustness of the evaluation, we conduct experiments with five repeated evaluations. \gptfourv with 8 frames as visual input achieves an average accuracy of 22.12\% with a deviation of 0.42\% on AutoEval-Video. Furthermore, the notable gap in performance between humans and VLMs suggests that AutoEval-Video reveals fundamental limitations of these models, independent of evaluation variability.

\subsection{Sustainability of AutoEval-Video}
We believe that AutoEval-Video is a sustainable benchmark, even when VLMs incorporate the videos in AutoEval-Video for training. We conduct experiment to verify this claim. Since human speech is commonly used as supervision in video-text pre-training, we extract human speech from these videos and add it to the input of the VLMs for re-evaluation. Even with ground truth text, \gptfourv only achieves 21.8\% accuracy, similar to its original 22.2\% accuracy over 8 frames. This suggests that training with the associated metadata will not improve the accuracy on our benchmark.

\subsection{Fine-Grained Assessment on AutoEval-Video}
In addition to allowing the automatic evaluator to make binary decisions, we also provide finer-grained assessments across three dimensions: language fluency, alignment to the question, and level of hallucination. The prompt used is as follows:

\noindent\begin{tikzpicture}[mybox/.style={rectangle, rounded corners, draw=black, fill=gray!10, thick, inner sep=10pt, align=justify, text width=0.9\linewidth}]
\node[mybox] {Moreover, there are three different aspects to consider when evaluating the responses. These aspects include:\newline
1. Validity of the answer (answers that are gibberish or exhibit excessive repetition of the same sentence structure, which defies language logic, are invalid)\newline
2. Whether the answer addresses the question (regardless of whether it’s correct or not)\newline
3. The presence of hallucinatory phenomena when compared to the video description (hallucination refers to activities or things in the answer that are not mentioned in the video description)\newline
At the end of your output result, please start a new line, and in the format of a dictionary, state your judgment on the above three aspects. For example, `{'text validity': "yes", 'adherence to questions': "no", 'hallucination detection': "yes"}` indicates that the response is valid but does not address the question, and there is a presence of hallucination.};
\end{tikzpicture}

To validate the fine-grained automatic evaluator, we randomly sampled 50 (video, question, response) triplets, comprising both model-generated and human-generated responses. On this test dataset, the fine-grained automatic evaluator aligns with human expert evaluation at a 94.0\% agreement rate.

\begin{table*}
    \centering
    \renewcommand{\arraystretch}{1.75}
    \setlength{\tabcolsep}{3pt}
    \resizebox{\linewidth}{!}{%
    \begin{tabular}{lccccccccc}
    \toprule
    \multirow{3}{*}{Model} &  \multicolumn{9}{c}{Skill} \\
    \cline{2-10}
        & Dynamic & State Transitions & Comparison & Reasoning with  & Explanatory & Predictive & \multirow{2}{*}{Description} & Counterfactual & Camera Movement  \\
        & Perception & Perception & Reasoning & External Knowledge & Reasoning& Reasoning &  & Reasoning& Perception\\
\midrule
\bliptwo & 0.0 & 0.0 & 0.0 & 0.0 & 1.5 & 0.0 & 0.0 & 0.0 & 0.0 \\
\instructblip & 6.5 & 12.5 & 21.1 & 6.3 & 7.6 & 0.0 & 0.0 & 26.3 & 0.0 \\
\qwenvl & 4.3 & 6.2 & 21.1 & 7.5 & 4.5 & 12.5 & 3.3 & 5.3 & 0.0 \\
\llava & 10.9 & 9.4 & 5.3 & 8.8 & 3.0 & 6.2 & 10.0 & 15.8 & 33.3 \\
\gptfourv & 28.3 & 25.0 & 10.5 & 30.0 & 22.7 & 18.8 & 20.0 & 21.1 & 0.0 \\
\videochatgpt & 2.2 & 6.2 & 15.8 & 11.2 & 4.5 & 3.1 & 0.0 & 15.8 & 0.0 \\
\videochat & 8.7 & 9.4 & 31.6 & 11.2 & 16.7 & 9.4 & 6.7 & 26.3 & 33.3 \\
\videollama & 4.3 & 6.2 & 21.1 & 13.8 & 10.6 & 12.5 & 3.3 & 26.3 & 0.0 \\
LLaMA-VID & 4.3 & 9.4 & 21.1 & 11.2 & 13.6 & 3.1 & 6.7 & 10.5 & 0.0 \\
VideoChat2 & 0.0 & 0.0 & 0.0 & 0.0 & 0.0 & 0.0 & 3.3 & 0.0 & 0.0 \\
Gemini-Ultra-Vision & 22.2 & 12.5 & 27.8 & 21.2 & 13.6 & 9.7 & 16.7 & 26.3 & 0.0 \\
\bottomrule
    \end{tabular}}
    \vspace{5pt}
    \caption{\bf Results of models utilizing 8-frame visual input on different skill dimensions in AutoEval-Video.}
    \label{evaluate result of 8 sampled frames on different dimensions}
\end{table*}

\begin{table*}
    \centering
    \renewcommand{\arraystretch}{1.75}
    \setlength{\tabcolsep}{3pt}
    \resizebox{\linewidth}{!}{%
    \begin{tabular}{lccccccccc}
    \toprule
    \multirow{3}{*}{Model} &  \multicolumn{9}{c}{Skill} \\
    \cline{2-10}
        & Dynamic & State Transitions & Comparison & Reasoning with  & Explanatory & Predictive & \multirow{2}{*}{Description} & Counterfactual & Camera Movement  \\
        & Perception & Perception & Reasoning & External Knowledge & Reasoning& Reasoning &  & Reasoning& Perception\\
\midrule
\bliptwo & 0.0 & 0.0 & 5.3 & 1.2 & 1.5 & 0.0 & 0.0 & 5.3 & 0.0 \\
\instructblip & 15.2 & 15.6 & 21.1 & 15.2 & 10.6 & 6.2 & 10.0 & 31.6 & 0.0 \\
\qwenvl & 10.9 & 3.1 & 21.1 & 10.0 & 9.1 & 6.2 & 6.7 & 26.3 & 0.0 \\
\llava & 8.7 & 6.2 & 5.3 & 10.0 & 4.5 & 0.0 & 6.7 & 31.6 & 33.3 \\
\gptfourv & 37.0 & 34.4 & 31.6 & 30.0 & 28.8 & 31.2 & 33.3 & 36.8 & 33.3 \\
\videochatgpt & 4.4 & 0.0 & 10.5 & 5.0 & 7.6 & 3.1 & 0.0 & 10.5 & 0.0 \\
\videochat & 8.7 & 15.6 & 21.1 & 8.8 & 10.6 & 3.1 & 10.0 & 21.1 & 0.0 \\
\videollama & 6.5 & 6.2 & 21.1 & 10.0 & 6.1 & 0.0 & 3.3 & 31.6 & 0.0 \\
LLaMA-VID & 8.7 & 12.5 & 15.8 & 8.8 & 3.0 & 6.2 & 6.7 & 5.3 & 33.3 \\
VideoChat2 & 2.2 & 0.0 & 0.0 & 0.0 & 0.0 & 0.0 & 3.3 & 0.0 & 0.0 \\
Gemini-Ultra-Vision & 35.6 & 21.9 & 31.6 & 15.0 & 24.2 & 5.6 & 3.3 & 26.3 & 0.0 \\
\bottomrule
    \end{tabular}}
    \vspace{5pt}
    
    \caption{\bf Results of models using 16-frame visual input on different skill dimensions in AutoEval-Video.}
    \label{evaluate result of 16 sampled frames on different dimensions}
\end{table*}

\begin{table*}[htbp]
\vspace{-10pt}
    \centering
    \renewcommand{\arraystretch}{1.25}
    \setlength{\tabcolsep}{3pt}
    \resizebox{\linewidth}{!}{%
    \begin{tabular}{lccccccccc}
    \toprule
    \multirow{3}{*}{Skill} 
        & Dynamic & State Transitions & Comparison & Reasoning with  & Explanatory & Predictive & \multirow{2}{*}{Description} & Counterfactual & Camera Movement  \\
        & Perception & Perception & Reasoning & External Knowledge & Reasoning& Reasoning &  & Reasoning& Perception\\
\midrule
Human & 82.6 (+54.3) & 68.8 (+43.8) & 80.70 (+70.2) & 70.0 (+40.0) & 82.3 (+59.6) & 78.1 (+59.3) & 54.4 (+34.4) & 82.5 (+61.4) & 44.4 (+44.4)\\
\bottomrule
    \end{tabular}}
\vspace{10pt}
    \caption{Results of humans on different skill dimensions, including the improvements compared to GPT-4V.}
    \label{human-nine-dim}
\end{table*}

\subsection{Assess the auto evaluator on binary test set}
We also construct a test dataset of (question, model prediction, judgment) triplets, where judgment is a ground truth binary indicator of the model prediction's correctness, based on the consensus of three human experts. The dataset is balanced, with half \texttt{true} and half \texttt{false} judgments. On this set, the automatic evaluator achieves a 94.79\% agreement rate with the ground truth.

\section{Case Study}\label{case-study}
\subsection{Hard Cases for Open-Sourced Models}\label{models-hard-cases}
In this subsection, we provide examples of three typical categories of bad cases generated by open-sourced models: hallucination, alignment, and weak LLM. These examples can be found in ~\cref{hallucination}, ~\cref{instructions_follow_fail}, and ~\cref{confusion and duplicate response}, respectively.

\subsection{Case Study for \videochat}\label{VideoChat-Case}

Among all open-sourced models, \videochat achieves the highest score. Furthermore, \gptfourv accurately answers just 32.7\% of cases where \videochat is correct. Examining their responses on \autoevalvideo reveals that \videochat outperforms \gptfourv in three areas: 1) traffic-related tasks; 2) comparisons involving dynamic objects; and 3) generating video-specific responses rather than general, vague answers. See \cref{videoChat-GPT-4V} for relevant examples.

\subsection{16-Frame vs. 8-Frame}
\label{16and8frames}

By increasing the visual input to 16 frames, \gptfourv achieves significant improvement compared to the 8-frame setting. In \cref{16-frame}, we provide examples of instances where questions cannot be accurately answered with an 8-frame input, but can be resolved correctly when utilizing a 16-frame input.

\subsection{Case Study for \gptfourv}\label{case-study-gpt-4v}

Since \gptfourv is state-of-the-art for now, we present a comprehensive error analysis for it on \autoevalvideo. In \cref{gpt-4v-badcase-dis}, we categorize the challenging cases encountered by \gptfourv for various reasons. The primary cause for the difficulty is related to the insufficient number of frames. Another significant issue is its tendency to produce overly general responses, as demonstrated in \cref{generalized-resp-tendency}. Furthermore, \gptfourv struggles with temporal and dynamic comprehension, including understanding of continuous visual features (\cref{continuous-features}), understanding across multiple frames (\cref{dynamic-perception} and \cref{cross-shot}), and moving objects (\cref{comparison-reasoning}).

\begin{table}[thbp]
\centering
\setlength{\tabcolsep}{12pt}
    \renewcommand{\arraystretch}{1.2}
  \resizebox{0.8\linewidth}{!}{%
\begin{tabular}{lc}
\toprule
 Bad Case Category & Proportion(\%)\\
  \midrule
Insufficient Frames & 27.92\\
Overly General Answers & 19.48\\
Temporal and Dynamic Understanding & 16.88\\
Reasoning & 9.74\\
Hallucination & 7.14\\
Object Perception & 5.19\\
Reject to Answer & 5.19\\
Others & 8.44\\
\bottomrule
\end{tabular}
}
\vspace{5pt}
\caption{\bf The distribution of bad cases for \gptfourv on \autoevalvideo.}
\vspace{-10pt}
\label{gpt-4v-badcase-dis}
\end{table}

\begin{table*}[bp]
\centering
\renewcommand{\arraystretch}{1.25}
\resizebox{\linewidth}{!}{%
\begin{tabular}{p{18cm}}

    \toprule
  {\bf Question:} What occurred in the cup? \hfill [{\bf ID:}169] \\
  \includegraphics[width=1\linewidth]{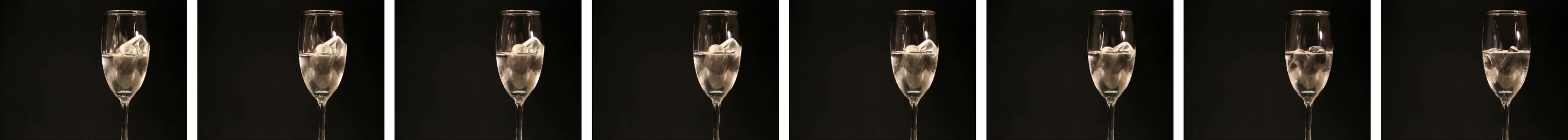} \\
  \midrule
  {\bf Sample Answer:} The ice melted. \\
  \midrule
  {\bf \llava:} In the cup, there is a small amount of water that has begun \textcolor{red}{\bf overflowing}, and at the same moment, the cup contains only a drop of water.\\
  \midrule
  {\bf \instructblip:} The liquid in the cup was \textcolor{red}{\bf poured from the left side to the right side}. \\
  
    \bottomrule
    \toprule
  
  {\bf Question:} Describe the two extra features of this vacuum cleaner sequentially. \hfill [{\bf ID:}294] \\
  \includegraphics[width=1\linewidth]{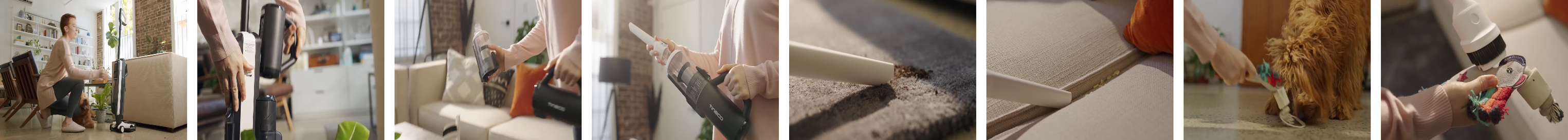} \\
  \midrule
  {\bf Sample Answer:} The vacuum cleaner can be paired with two unique attachments, each serving a distinct function: first, when paired with the long nozzle attachment, it can clean up small debris on carpets and sofas; second, when paired with the brush attachment, it can remove dust from plush toys. \\
  \midrule
    {\bf \qwenvl:} The vacuum cleaner has a powerful motor, which helps in cleaning the carpet efficiently and removing dirt, dust and debris. \textcolor{red}{\bf It also has a HEPA filter that captures allergens and pollen from the air, making it an ideal choice for people with allergies}.\\
  \midrule
    {\bf \llava:} This vacuum cleaner has two additional functions, as shown in the image. The first function is a mop cleaning tool, which helps to clean the floors. A woman in the scene is seen using a mop with the vacuum cleaner attached, allowing her to both vacuum and mop simultaneously. \textcolor{red}{\bf The second function is a shoe cleaner}, which adds additional versatility for cleaning various items and surfaces, including shoes. The vacuum cleaner is being used to clean \textcolor{red}{\bf shoes in the picture}. These two functions make the vacuum cleaner a more versatile and comprehensive appliance for cleaning purposes.\\
  
    \bottomrule
    \toprule
  
  {\bf Question:} What is the shortest time duration mentioned in the video? \hfill [{\bf ID:}14] \\
  \includegraphics[width=1\linewidth]{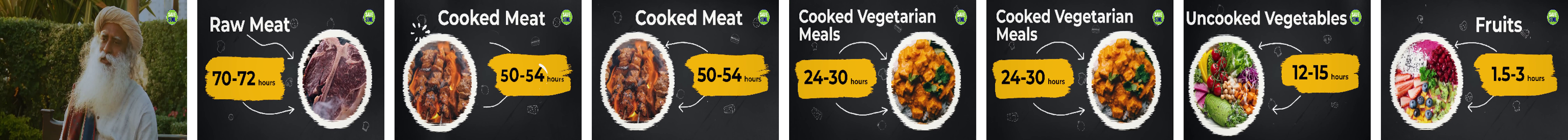} \\
  \midrule
  {\bf Sample Answer:} 1.5 hours. \\
  \midrule
    {\bf \qwenvl:} \textcolor{red}{\bf 15-30 min}.\\
  \midrule
    {\bf \llava:} The shortest time duration mentioned in the video is \textcolor{red}{\bf 15 minutes}.\\
  \midrule
    {\bf \videochat:} The shortest time duration mentioned in the video is \textcolor{red}{\bf 12 minutes}.\\
  \midrule
    {\bf \videochatgpt:} The shortest time duration mentioned in the video is \textcolor{red}{\bf 15 minutes}.\\
  \midrule
    {\bf \instructblip:} \textcolor{red}{\bf 10 seconds}.\\
  \midrule
    {\bf \videollama:} The shortest time duration mentioned in the video is \textcolor{red}{\bf 21.8 seconds}.\\
  \bottomrule
\end{tabular}
}
\vspace{5pt}
\caption{\bf Examples of model responses on \autoevalvideo with hallucination.}
\label{hallucination}
\end{table*}

\begin{table*}[tbp]
\centering
\renewcommand{\arraystretch}{1.25}
\resizebox{0.99\linewidth}{!}{%
\begin{tabular}{p{18cm}}
    \toprule
  {\bf Question:} \textcolor{red}{\bf Who} are the potatoes in the bowl intended for? \hfill [{\bf ID:}208] \\
  \includegraphics[width=1\linewidth]{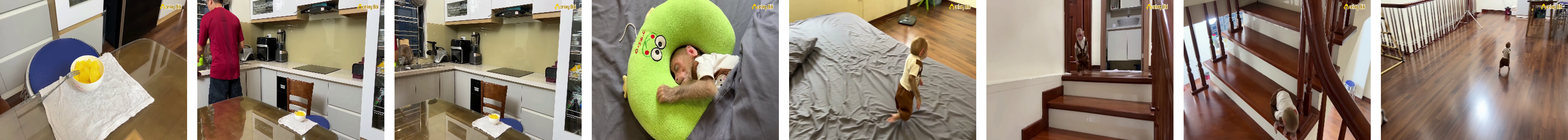} \\
  \midrule
  {\bf Sample Answer:} The monkey. \\
  \midrule
  {\bf \videollama:} In the first scene, we see a young man holding a bowl of potatoes on a wooden table in a kitchen. The table has a white countertop and a wooden floor. The man is wearing a red shirt and holding the bowl of potatoes with his right hand. We can also see a silver spoon in the bowl.\\
  \midrule
  {\bf \instructblip:} mashed.\\
    \bottomrule
    \toprule
  {\bf Question:} The video features three animals: fish, birds, and cats. Please sequence them in the order of their appearance. \hfill [{\bf ID:}33] \\
  \includegraphics[width=1\linewidth]{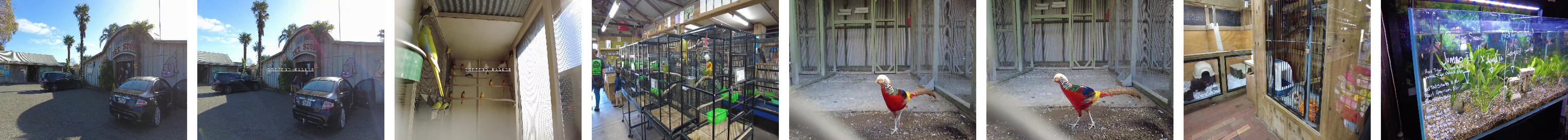} \\
  \midrule
  {\bf Sample Answer:} Birds, cats, and fish. \\
  \midrule
  {\bf \videochatgpt:} We can't provide an exact sequence of the animals based on the image description alone. However, what is evident is that there are three different animals present in the video. Since the video does not provide a clear order, we do not have any further information on the animals' appearance in the video.\\
  \midrule
  {\bf \bliptwo:} The sequence should be as follows: fish, birds, cats, fish, birds, cats, fish, birds, cats, fish, birds...\\
  \midrule
  {\bf \llava:} Fish, birds, cats, birds, cats, fish.\\
    \bottomrule
    \toprule
  {\bf Question:} What is the first thing the person cuts \textcolor{red}{ \bf after slicing this red pepper?} \hfill [{\bf ID:}74]\\
  \includegraphics[width=1\linewidth]{fig/74.png} \\
  \midrule
  {\bf Sample Answer:} The onion. \\
  \midrule
  {\bf \videochat:} The person firstly \textcolor{red}{\bf cuts the red pepper into slices}.\\
  \midrule
  {\bf \videollama:} The person firstly \textcolor{red}{\bf cuts the red pepper with a knife on a cutting board}.\\
    \bottomrule
  
\end{tabular}}
\vspace{5pt}
\caption{\bf Examples of model responses on \autoevalvideo that do not adhere to the given questions.}
\label{instructions_follow_fail}
\end{table*}

\begin{table*}[tbp]
\centering
\renewcommand{\arraystretch}{1.25}
\resizebox{0.99\linewidth}{!}{%
\begin{tabular}{p{18cm}}

    \toprule
  {\bf Question:} Describe the two extra features of this vacuum cleaner sequentially. \hfill [{\bf ID:}294] \\
  \includegraphics[width=1\linewidth]{fig/294.png} \\
  \midrule
  {\bf Sample Answer:} The vacuum cleaner can be paired with two unique attachments, each serving a distinct function: first, when paired with the long nozzle attachment, it can clean up small debris on carpets and sofas; second, when paired with the brush attachment, it can remove dust from plush toys. \\
  \midrule
  {\bf \videollama:} First, \textcolor{red}{\bf we see a woman holding a vacuum cleaner and cleaning the floor in her living room}. Next, \textcolor{red}{\bf we see a woman holding a vacuum cleaner and cleaning the floor in her living room}. Then, \textcolor{red}{\bf we see a woman holding a vacuum cleaner and cleaning the floor in her living room}. After that, we see a woman holding a vacuum cleaner and cleaning the floor in her living room. Next, we see a woman holding a vacuum cleaner and cleaning the floor in her living room. Then, we see a woman holding a vacuum cleaner and cleaning the floor in her living room...\\
  \midrule
  {\bf \llava:} The two extra features of this vacuum cleaner include \textcolor{red}{\bf two people using the tool in sequence and the inclusion of items such as a sandwich and a dog amid the cleaning process}.  The image displays a montage of the tool in action, capturing diverse aspects like individuals working with the vacuum side by side, which suggests a collaborative approach to cleaning. Additionally, the presence of a sandwich and a dog also suggests that the image might be representing a break during cleaning, indicating that the workers or the creator of this imagery aimed to depict everyday life alongside these tasks.\\
    \bottomrule
    \toprule
  {\bf Question:} What action did the individual take after peeling these potatoes? \hfill [{\bf ID:}41]\\
  \includegraphics[width=1\linewidth]{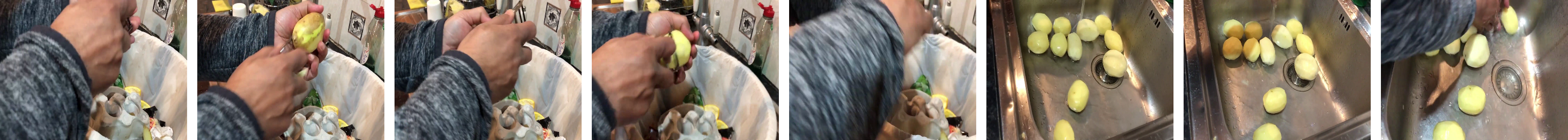} \\
  \midrule
  {\bf Sample Answer:} Wash the potatoes. \\
  \midrule
  {\bf \bliptwo:} $|$ $|$ $|$ $|$ $|$ $|$ $|$ $|$ $|$ $|$ $|$ $|$ $|$ $|$ $|$ $|$ $|$ $|$ $|$ $|$ $|$ $|$ $|$ $|$ ...\\
    \bottomrule
    \toprule
  {\bf Question:} What is the shortest time duration mentioned in the video? \hfill [{\bf ID:}33] \\
  \includegraphics[width=1\linewidth]{fig/33.png} \\
  \midrule
  {\bf Sample Answer:} Birds, cats, and fish. \\
  \midrule
    {\bf \qwenvl:} (ref)(fish)(ref)(box)(760,231)\\
\bottomrule
\end{tabular}}
\vspace{5pt}
\caption{\bf Examples of model responses on \autoevalvideo exhibiting limited language generation capabilities.}
\label{confusion and duplicate response}
\end{table*}

\begin{table*}[tbp]
\centering
\renewcommand{\arraystretch}{1.25}
\resizebox{0.99\linewidth}{!}{%
\begin{tabular}{p{18cm}}

    \toprule
  {\bf Question:} What is the correct action for this car that's waiting to turn left? \hfill [{\bf ID:}105] \\
  \includegraphics[width=1\linewidth]{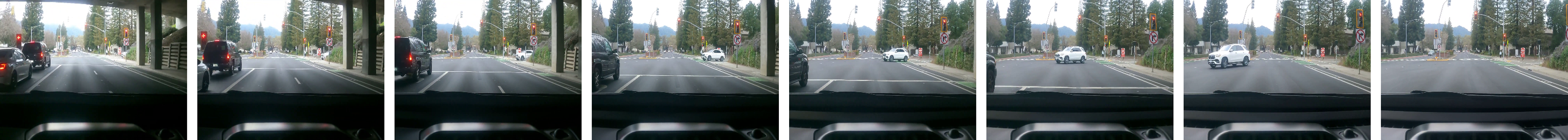} \\
  \midrule
  {\bf Sample Answer:} It should wait for the green light. \\
  \midrule
  {\bf VideoChat:} The car waiting to turn left should wait until the traffic light turns green before turning.\\
  \midrule
  {\bf GPT-4V:} The car waiting to turn left should yield to oncoming traffic. Once it is safe and there are no vehicles approaching from the opposite direction or if there is a sufficient gap between vehicles, the car can then proceed with the left turn. If there is a dedicated left-turn signal or if the light turns yellow and oncoming traffic stops, the car can also proceed with the turn. It's always crucial to ensure the way is clear and safe before making a turn. Safety first! \\
    \bottomrule
    \toprule
  
  {\bf Question:} In the video, which car is moving at a slower pace? \hfill [{\bf ID:}191] \\
  \includegraphics[width=1\linewidth]{fig/191.png} \\
  \midrule
  {\bf Sample Answer:} The police car. \\
  \midrule
  {\bf VideoChat:} In the video, the police car is going slower than the ambulance.\\
  \midrule
  {\bf GPT-4V:} The image appears to be a still frame or a sequence of similar frames, so it's not possible to determine the speed of the cars based on this image alone. To accurately determine which car is going slower, you would need to see a video or have more context. \\
    \bottomrule
    \toprule
  {\bf Question:} What is the relationship between these individuals? \hfill [{\bf ID:}26] \\
  \includegraphics[width=1\linewidth]{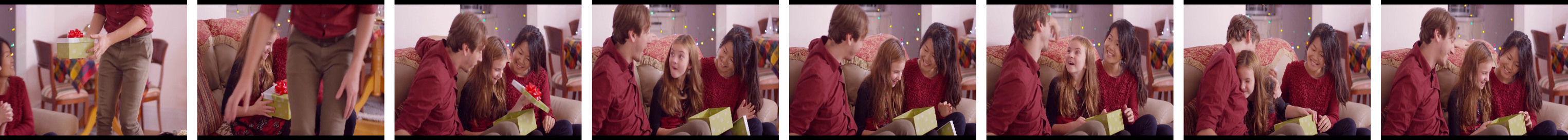} \\
  \midrule
  {\bf Sample Answer:} They are family. \\
  \midrule
  {\bf VideoChat:} It appears that the people in the video are a family, possibly a mother, father, and daughter.\\
  \midrule
  {\bf GPT-4V:} The image appears to depict multiple individuals in a home setting, likely engaging in a shared activity or celebration. The exact relationship between the individuals cannot be definitively determined based on the image alone. They could be family members, friends, or any other type of acquaintances. However, the intimate and friendly nature of their interactions suggests they have a close relationship, potentially as family or good friends. \\
\bottomrule
  
\end{tabular}}
\vspace{5pt}
\caption{{\bf \videochat vs. \gptfourv}: \videochat outperforms \gptfourv on three domains, traffic-related tasks (instance 105), comparison between dynamic objects (instance 191), and generating video-specific responses rather than general answers (instance 26).}
\label{videoChat-GPT-4V}
\end{table*}
\begin{table*}[tbp]
\centering
\vspace{-15pt}
\resizebox{0.95\linewidth}{!}{%
\begin{tabular}{p{18cm}}

    \toprule
  {\bf Question:} What action did the individual take towards the flower after he opened the book? \hfill [{\bf ID:}49] \\
  \midrule
  8 frames as visual input:\\
  \includegraphics[width=1\linewidth]{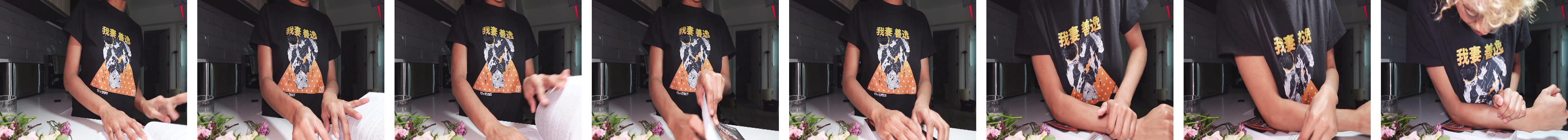} \\
  \midrule
  16 frames as visual input:\\
  \includegraphics[width=1\linewidth]{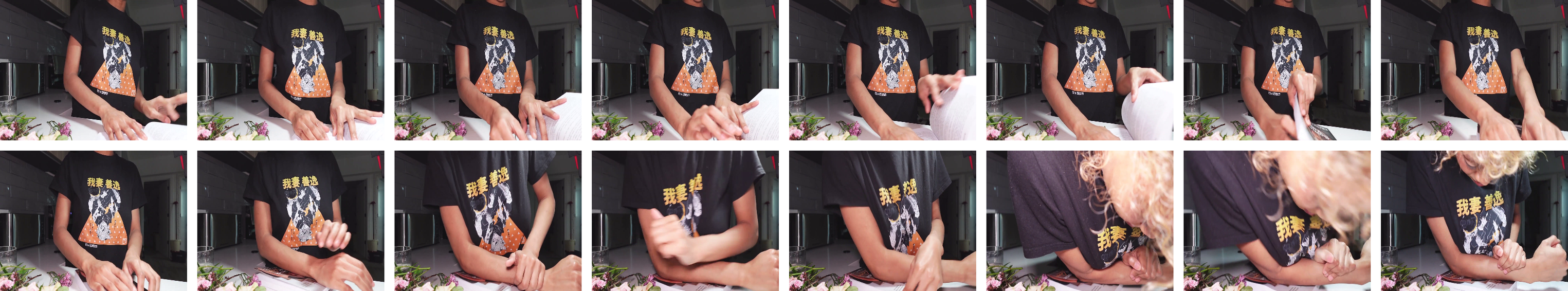} \\
  \midrule
  {\bf Sample Answer:} Put the flowers in a book and press the book. \\
  \midrule
  {\bf \gptfourv (8-frame):} The sequence of images shows an individual at a table with a book and a flower lying next to it.From left to right, the sequence depicts the person opening the book, turning pages, then picking up the flower and appearing to smell it after the book has been opened.The final image shows the person with their head bowed down towards the flower held in their hand, suggesting the action of smelling the flower.\\
  \midrule
  {\bf \gptfourv (16-frame):} The series of images depict a sequence of actions. The individual in the images appears to be engaging in a craft or a demonstration. After opening the book, \textcolor{blue}{\bf he have taken a flower, placed it within the pages of the book}, and then closed the book, likely to press the flower for preservation or as part of a craft project. This is a common method for pressing and drying flowers, which can later be used for art, decoration, or keepsakes.\\
    \bottomrule
    \toprule
  {\bf Question:} How did the direction of the car change? \hfill [{\bf ID:}61] \\
  \midrule
  8 frames as visual input:\\
  \includegraphics[width=1\linewidth]{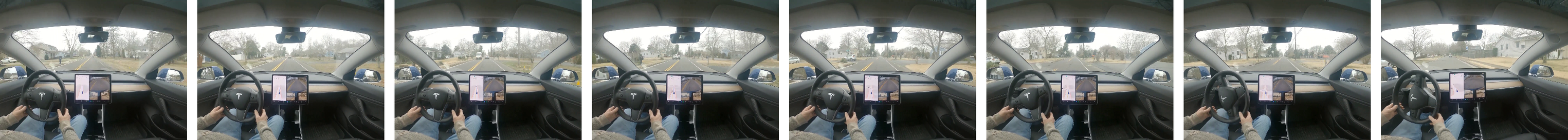} \\
  \midrule
  16 frames as vidual input:\\
  \includegraphics[width=1\linewidth]{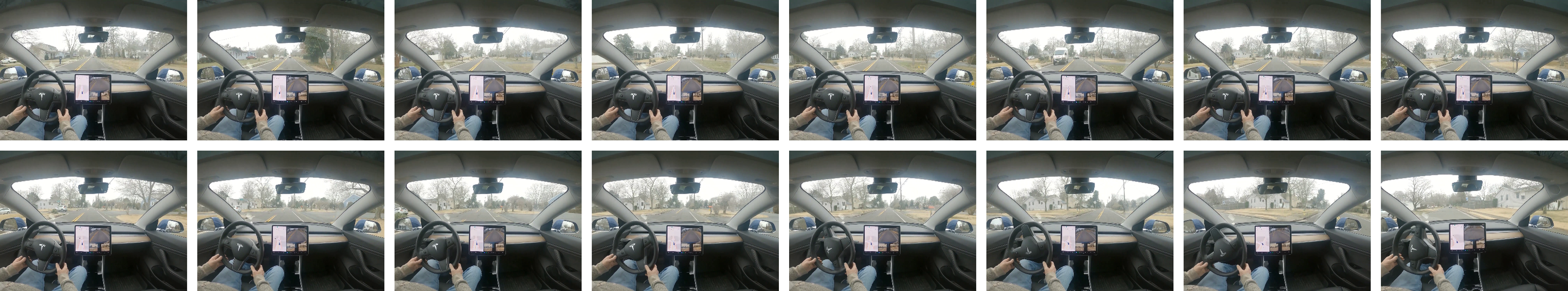} \\
  \midrule
  {\bf Sample Answer:} The car went straight and then turned left. \\
  \midrule
  {\bf \gptfourv (8-frame):} It appears you've provided a panoramic or 360-degree type image that has been stretched into a flat perspective, showing multiple repeated views of the inside of a car, most likely a Tesla considering the logo on the steering wheel and the characteristic dashboard design.However, due to the nature of the image, it's challenging to ascertain the direction of the car's movement with certainty. The panoramic stretching distorts the scene and repeats it, which does not provide a reliable sense of direction change.In typical images or video, one might look for changes in the car's orientation relative to the road or the movement of the steering wheel to determine a direction change. However, in this image, there's no clear evidence that can be used to determine how the car's direction may have changed. The image would need to be in a standard perspective to make such an assessment.\\
  \midrule
  {\bf \gptfourv (16-frame):} The series of images shows the interior view of a vehicle from the driver's perspective, specifically a Tesla, as evidenced by the logo on the steering wheel. Across the sequence of images, we can see the vehicle's progression as it makes a left turn:
1.Initial Frames (Top Row): The car starts off \textcolor{blue}{\bf going straight on a road}.
2.Middle Frames (Middle Rows): As we progress through the frames, we see that \textcolor{blue}{\bf the driver turns the steering wheel to the left}, initiating the turn.
3.Final Frames (Bottom Row): The car has completed much of the turn.The steering wheel is returning to the neutral position as the car aligns with the new direction of the road after the turn.
The environment outside the vehicle changes consistently with the progression of a left turn, starting with houses on the side and transitioning to them being visible through the rear as the turn progresses.The motion is smooth, indicating a continuous action, which suggests these frames capture a sequence from a video of a car turning left at an intersection or onto another road.\\
\bottomrule
  
\end{tabular}}
\vspace{5pt}
\caption{{\bf 16-Frame vs. 8-Frame}: Examples of instances in which GPT-4V correctly responds only when utilizing 16 frames as visual input.}
\label{16-frame}
\vspace{-20pt}
\end{table*}
\begin{table*}[tbp]
\centering
\vspace{-10pt}
\renewcommand{\arraystretch}{1.25}
\resizebox{0.95\linewidth}{!}{%
\begin{tabular}{p{18cm}}

    \toprule
  {\bf Question:} Why did the chicken turn into an egg? \hfill [{\bf ID:}36] \\
  \includegraphics[width=1\linewidth]{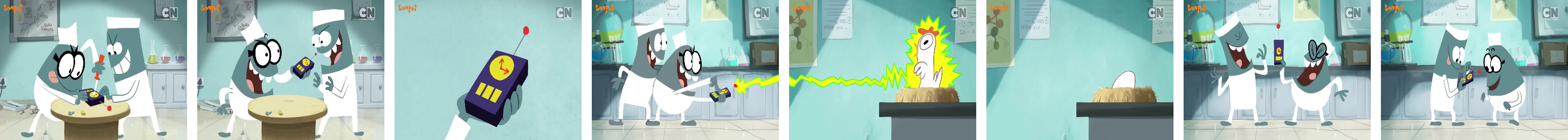} \\
  \midrule
  {\bf Sample Answer:} Because these two guys were shining a machine at the chicken. \\
  \midrule
  {\bf GPT-4V:} The images you provided appear to be from a cartoon, possibly referencing a comedic or fictional scenario. In cartoons and animated series, various humorous and impossible situations can be presented for comedic effect. The concept of a chicken turning into an egg might be a play on the age-old question: ``Which came first, the chicken or the egg?" Without specific context from the show, it's hard to pinpoint the exact reason, but it's likely intended as a humorous twist on that question. \\
    \bottomrule
    \toprule
  {\bf Question:} Why is the individual on the left heading in that particular direction? \hfill [{\bf ID:}266] \\
  \includegraphics[width=1\linewidth]{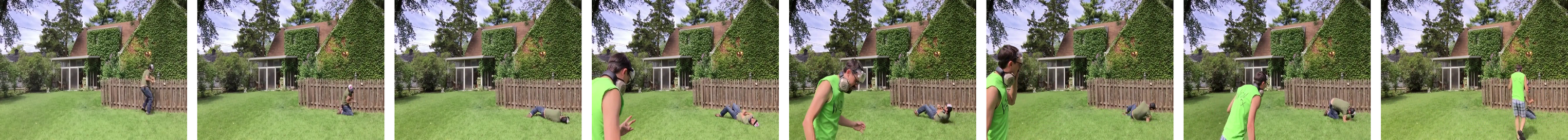} \\
  \midrule
  {\bf Sample Answer:} Because the person on the left wants to confirm that the person on the right is not injured. \\
  \midrule
  {\bf GPT-4V:} Based on the image, I can't determine the specific reason why the person on the left is walking in that direction. People might walk for various reasons, such as exploring, heading towards a destination, or simply enjoying the outdoors. Without additional context, it's hard to say for certain. \\
    \bottomrule
    \toprule
  {\bf Question:} Please provide step-by-step instructions on how to operate this coffee machine as demonstrated in the video. \hfill [{\bf ID:}197]\\
  \includegraphics[width=1\linewidth]{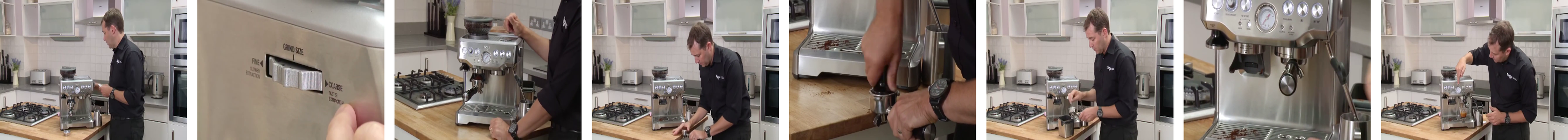} \\
  \midrule
  {\bf Sample Answer:} First, select the double filter just put that in the handle; Second, get our grind size towards the finer end; Third, as we've used the double filter, select the double filter size and activate the grinder, simply bump and release; Four, tap and get that coffee into the filter; Five, grab our tamp, give it a tap; Six, use razor scraper tool to scrape off the excess coffee to ensure that we've got our volumes of coffee right built for this machine; Seven, insert the handle and extract; Eight, place the cup just below the outlet. \\
  \midrule
  {\bf GPT-4V:} Certainly! Based on the series of images provided, here are the steps on how to use the coffee machine:
1. Preparation: Start by making sure the coffee machine is clean, plugged in, and turned on. Check if there's enough water in the reservoir.
2. Coffee Beans: Place the coffee beans in the grinder hopper on top of the machine.
3. Grinding: Adjust the grind setting to your preference. Some machines have multiple settings, ranging from fine to coarse. Once set, activate the grinder to grind the beans. The ground coffee will collect in a portafilter or container.
4. Tamping: Once you have enough ground coffee in the portafilter, use a tamper to press down the coffee grounds. This ensures even water distribution during brewing and a consistent espresso shot. Apply firm and even pressure while tamping.
5. Inserting the Portafilter: Attach the portafilter to the machine by slotting it into the group head. Ensure it's securely locked in place.
6. Brewing: Press the appropriate button or turn the lever to start the brewing process. Hot water will pass through the coffee grounds, extracting flavors and pouring the coffee into the cup below.
7. Steam Wand (If Applicable): If you want to froth milk for a cappuccino or latte, fill a stainless steel jug with cold milk. Place the steam wand into the milk and activate it. This will introduce steam into the milk, causing it to froth and increase in volume. Move the jug in an up and down motion to get even frothing.
8. Pouring Serving: Once you have your brewed coffee and frothed milk (if applicable), pour them into your cup to create your desired coffee drink.
9. Cleaning: After use, ensure you clean the machine parts, such as the portafilter, tamper, and steam wand, to maintain the machine and ensure the best coffee taste.
Remember, it's always a good idea to refer to the machine's user manual for specific instructions and maintenance tips. \\
\bottomrule
  
\end{tabular}}
\vspace{5pt}
\caption{\bf Examples of incorrect answers from GPT-4V demonstrate its tendency to provide answers in a general context rather than specifically addressing the content of the video.}
\label{generalized-resp-tendency}
\vspace{-10pt}
\end{table*}

\begin{table*}[tbp]
\centering
\renewcommand{\arraystretch}{1.25}
\resizebox{0.99\linewidth}{!}{%
\begin{tabular}{p{18cm}}
    \toprule
  {\bf Question:} When I am at the entrance, how can I reach the balcony that has three decorations on the table? \hfill [{\bf ID:}200] \\
  \includegraphics[width=1\linewidth]{fig/200.png} \\
  \midrule
{\bf Sample Answer:} You can move like the track of video camera, first turn left to the living room after entering the door, and then turn right. At this time, you are facing the destination balcony. You only need to go straight and open the balcony door to go out. \\
  \midrule
  {\bf \gptfourv:} Based on the images provided, it appears that you would enter through the main entrance, proceed through the living area, and then make your way to the open space that has sliding glass doors or large windows. This area seems to lead to the balcony with a seating arrangement and three decorations on the table. Ensure to look for the large sliding doors or windows to access the balcony. \\
    \bottomrule
    \toprule
  {\bf Question:} What is depicted in the video that does not occur in actuality? \hfill [{\bf ID:}305] \\
  \includegraphics[width=1\linewidth]{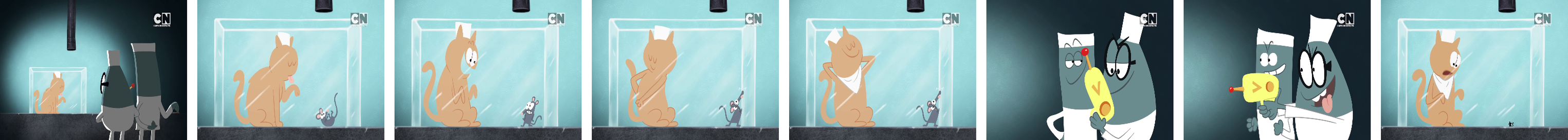} \\
  \midrule
  {\bf Sample Answer:} Mice can become smaller in an instant. \\
  \midrule
  {\bf \gptfourv:} The images depict a series of animated scenes involving cartoon characters. In these scenes, the actions defy real-world physics and logic. Specifically, a cat seems to be imitating the behavior of a fish in an aquarium, including blowing bubbles. This is unrealistic because cats cannot breathe underwater or blow bubbles like fish. Additionally, in the last image, the characters appear to be using a device to zap the cat, transforming its behavior or state, which is also fictional and not possible in reality. Such exaggerated and impossible actions are typical of cartoons for comedic or dramatic effect. \\
    \bottomrule
    \toprule
  {\bf Question:} How many characters from the blue team were killed in this video? \hfill [{\bf ID:}56] \\
  \includegraphics[width=1\linewidth]{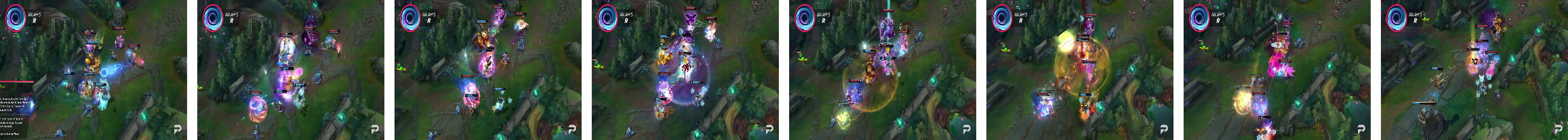} \\
  \midrule
{\bf Sample Answer:} Three. \\
  \midrule
  {\bf \gptfourv:} Based on the series of images provided, it looks like you're showing a sequence from a game of ``League of Legends" or a similar MOBA game. From the sequence, it appears there are several characters engaged in a team fight.
To determine how many characters on the blue team died, I'll look for blue character icons or health bars disappearing or turning gray.
Upon inspection, I couldn't identify any clear instances where a blue character's icon disappeared or their health bar turned gray. However, given the limited visibility and the sequential nature of the images, it's possible that some information might be missing or not clearly visible.
From the images provided, it seems that no characters from the blue team have died. However, for a more accurate count, it would be better to view the full video or provide more context about the game's events. \\
    \bottomrule
  
\end{tabular}}
\vspace{5pt}
\caption{{\bf Examples of tasks requiring the observation of continuous frames in AutoEval-Video.} For such instances, GPT-4V generates incorrect responses.}
\label{continuous-features}
\end{table*}

\begin{table*}[tbp]
\centering
\renewcommand{\arraystretch}{1.25}
\resizebox{0.99\linewidth}{!}{%
\begin{tabular}{p{18cm}}

    \toprule
  {\bf Question:} In the video, how many times do the two people pass the ball in total? \hfill [{\bf ID:}34] \\
  \includegraphics[width=1\linewidth]{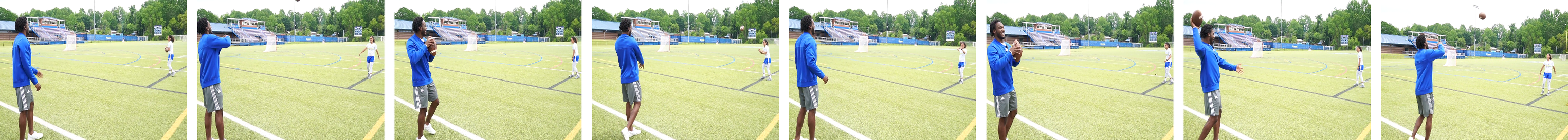} \\
  \midrule
{\bf Sample Answer:} Pass five times in total. \\
  \midrule
  {\bf \gptfourv:} The image you provided appears to be a sequence of still frames and \textcolor{red}{\bf does not contain any motion} that would allow me to count ball passes. Can you provide more details or ask another question? \\

    \bottomrule
    \toprule
  {\bf Question:} What actions did the individual perform between tossing the phone and picking up the book? \hfill [{\bf ID:}109] \\
  \includegraphics[width=1\linewidth]{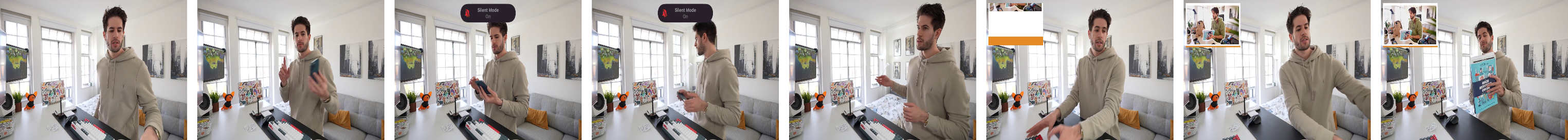} \\
  \midrule
  {\bf Sample Answer:} The person \textcolor{blue}{\bf moved the keyboard away}. \\
  \midrule
  {\bf \gptfourv:} The person \textcolor{red}{\bf waved his hand} and then reached out to pick up the book. \\

    \bottomrule
    \toprule
  {\bf Question:} Detail the individual's actions sequentially. \hfill [{\bf ID:}116] \\
  \includegraphics[width=1\linewidth]{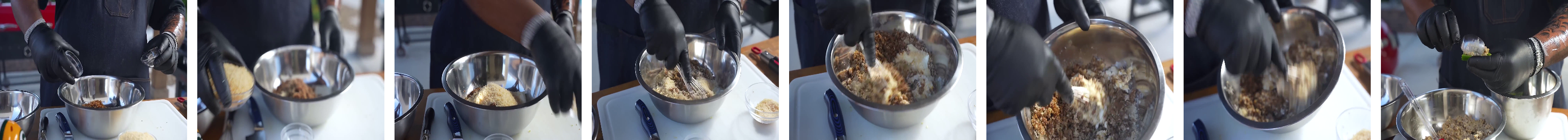} \\
  \midrule
{\bf Sample Answer:} First, sprinkle the seasoning on the chopped pork sausage; Second, sprinkle the cheese on the chopped pork sausage; Third, stir the above well; Fourth, place the stirred ingredients on the cut pepper. \\
  \midrule
  {\bf \gptfourv:} The sequence of images depicts a person preparing a dish. Here's the order of the steps:
1. The person is holding a bowl filled with some ingredients.
2. They begin \textcolor{red}{\bf mixing} the contents of the bowl.
3. The \textcolor{red}{\bf mixing} continues as the contents appear to blend together.
4. The person is still \textcolor{red}{\bf mixing}, ensuring the ingredients are well-combined.
5. The \textcolor{red}{\bf mix} seems to have a consistent texture now.
6. Finally, the person takes a small amount of the mixture, possibly for tasting or shaping.
Note: This description is based on the images provided and does not make any assumptions about the specific dish being prepared or the person's intentions. \\
\bottomrule

\end{tabular}}
\vspace{5pt}
\caption{\bf Examples of responses generated by GPT-4V that describe individual frames without effectively synthesizing information across multiple frames.}
\label{dynamic-perception}
\end{table*}

\begin{table*}[tbp]
\centering
\renewcommand{\arraystretch}{1.25}
\resizebox{0.99\linewidth}{!}{%
\begin{tabular}{p{18cm}}

    \toprule
  {\bf Question:} Why are the eyes of this orange animal wide open? \hfill [{\bf ID:}112] \\
  \includegraphics[width=1\linewidth]{fig/112.png} \\
  \midrule
{\bf Sample Answer:} Because the animal sees a train approaching the white animal on the tracks.\\
  \midrule
  {\bf \gptfourv:} The orange animal appears to be a stylized or animated character, possibly showing surprise, alarm, or excitement. Wide open eyes in animation are often used to convey strong emotions or reactions. This could be in response to something happening in the scene or to depict a particular personality trait of the character. Without additional context, it's hard to specify exactly why, but wide eyes are a common visual cue for heightened emotions in animation. \\

    \bottomrule
    \toprule
  
  {\bf Question:}  Which dish did these two individuals consume first? \hfill [{\bf ID:}117] \\
  \includegraphics[width=1\linewidth]{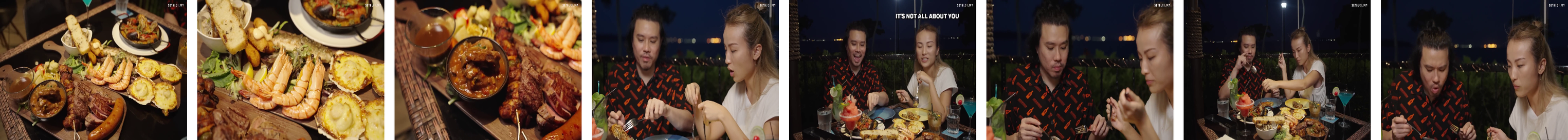} \\
  \midrule
  {\bf Sample Answer:} Paella. \\
  \midrule
  {\bf \gptfourv:} Based on the image, it's challenging to definitively determine which dish the two people ate first. Without the context of a video or sequence of images, I can't provide a definitive answer. However, you can look at the image to check which dishes appear to have been started or have portions taken out of them. \\

    \bottomrule
    \toprule
  
  {\bf Question:} What this spider will do next? \hfill [{\bf ID:}325] \\
  \includegraphics[width=1\linewidth]{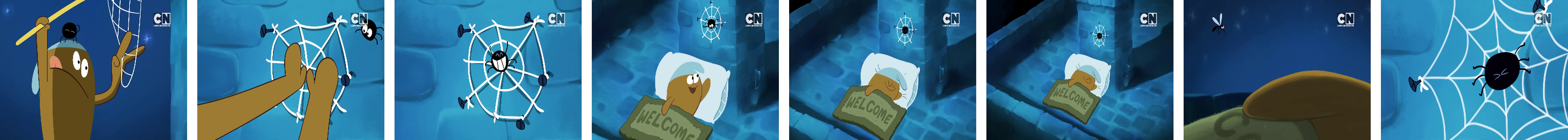} \\
  \midrule
{\bf Sample Answer:} Prevents mosquitoes from sucking blood. \\
  \midrule
  {\bf \gptfourv:} It seems like the spider has created a web and there's a ``WELCOME" sign placed on the web. Given the playful and mischievous nature often associated with animated characters like these, the spider might be planning a prank or trying to catch something in its web. However, without more context from the show or episode, it's just a fun speculation! \\
\bottomrule
  
\end{tabular}}
\vspace{5pt}
\caption{\bf Examples of incorrect responses by GPT-4V to video-questions related to cross-shot understanding in AutoEval-Video.}
\label{cross-shot}
\end{table*}

\begin{table*}[tbp]
\centering
\renewcommand{\arraystretch}{1.25}
\resizebox{0.99\linewidth}{!}{%
\begin{tabular}{p{18cm}}

    \toprule
  {\bf Question:} Discuss the changes in Super Mario's ranking in this video. \hfill [{\bf ID:}192] \\
  \includegraphics[width=1\linewidth]{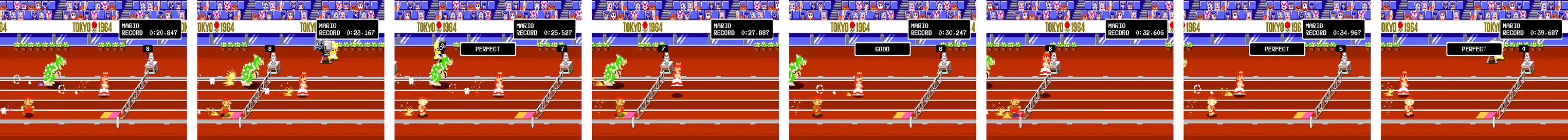} \\
  \midrule
{\bf Sample Answer:} Super Mario ran from last to first. \\
  \midrule
  {\bf \gptfourv:} The image provided seems to be a sequence or snapshot from a game that features Super Mario participating in a race, likely a reference to the ``Mario Sonic at the Olympic Games" series or something similar.
  From what can be observed:
1. In the first segment, Mario holds the 1st place.
2. In the second segment, Mario maintains his lead, still in 1st place.
3. By the third segment, Mario appears to have fallen slightly behind, taking the 2nd position.
4. In the fourth segment, Mario manages to regain his lead and is back to the 1st position.
5. In the fifth segment, Mario continues to hold onto the 1st position.
6. In the last segment, Mario still remains in the 1st position.
Overall, it's evident that Mario held a strong position throughout the race, primarily staying in 1st place. He did face some competition in the third segment, where he was temporarily overtaken but quickly managed to reclaim his lead. This shows Mario's persistence and determination in the race! \\
    \bottomrule
    \toprule
  {\bf Question:} In this competition, there are four participants, each wearing white, yellow, red, and blue helmets respectively. Which competitor outshines the rest? \hfill [{\bf ID:}247] \\
  \includegraphics[width=1\linewidth]{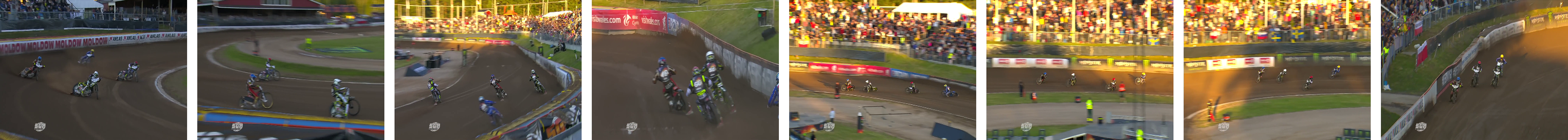} \\
  \midrule
{\bf Sample Answer:} Contestant with blue helmet. \\
  \midrule
  {\bf \gptfourv:} From the sequence of images provided, it appears that the rider with the white helmet is ahead of the riders with the yellow and red helmets in most of the frames. Therefore, the player wearing the white helmet outperformed the two opponents wearing yellow and red helmets. \\
    \bottomrule
    \toprule

{\bf Question:} In the video, there are two cars with different colors, one being red and the other blue. Which car is moving faster? \hfill [{\bf ID:}122]\\
  \includegraphics[width=1\linewidth]{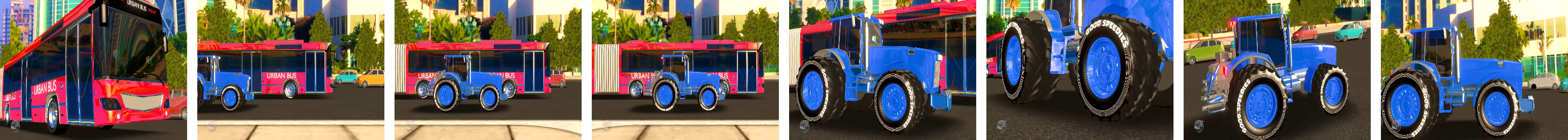} \\
  \midrule
  {\bf Sample Answer:} The blue car is faster. \\
  \midrule
  {\bf \gptfourv:} Based on the provided image, it's not possible to determine which car (or bus and tractor, in this case) is faster. The image is static and does not provide any information about their speeds. You would need a video or additional data to determine which one is faster. \\
\bottomrule
  
\end{tabular}
}
\vspace{5pt}
\caption{\bf Examples related to comparison reasoning in AutoEval-Video, for which GPT-4V is unable to generate correct responses.}
\label{comparison-reasoning}
\end{table*}

\end{document}